\documentclass[twoside]{article}

\usepackage[utf8]{inputenc} % allow utf-8 input
\usepackage[T1]{fontenc}
\usepackage{tgtermes}
\usepackage{amssymb}
\usepackage{amsfonts}

\usepackage{tgtermes}
\usepackage{amsmath}
\usepackage{scalefnt,letltxmacro}
\LetLtxMacro{\oldtextsc}{\textsc}
\renewcommand{\textsc}[1]{\oldtextsc{\scalefont{1.10}#1}}
\usepackage[scaled=0.92]{PTSans}

\usepackage[usenames,dvipsnames]{xcolor}
\definecolor{shadecolor}{gray}{0.9}

\usepackage[parfill]{parskip}
\usepackage{afterpage}
\usepackage{framed}
\usepackage{nicefrac}
\usepackage{paralist}
\usepackage[compact]{titlesec}
\titlespacing*{\section}{0pt}{3pt}{1ex}
\titlespacing{\subsection}{0pt}{1ex}{0ex}
\titlespacing{\subsubsection}{0pt}{0ex}{0ex}
\usepackage[colorinlistoftodos,
           prependcaption,
           textsize=small,
           backgroundcolor=yellow,
           linecolor=lightgray,
           bordercolor=lightgray]{todonotes}

\usepackage{lineno}
\renewcommand\linenumberfont{\normalfont
                             \footnotesize
                             \sffamily
                             \color{SkyBlue}}
\usepackage{ragged2e}

\DeclareRobustCommand{\parhead}[1]{\textbf{#1}~}

\usepackage{graphicx}
\usepackage{float}
\usepackage[labelfont=it,labelsep=period]{caption}
\usepackage[format=hang]{subcaption}
\usepackage{booktabs}
\usepackage{arydshln} % Dashed lines

\usepackage{natbib}

\usepackage{algorithm}
\usepackage{algorithmic}
\usepackage{listings}
\usepackage{fancyvrb}
\fvset{fontsize=\normalsize}

\usepackage[colorlinks,linktoc=all]{hyperref}
\usepackage[all]{hypcap}
\hypersetup{citecolor=Violet}
\hypersetup{linkcolor=black}
\hypersetup{urlcolor=MidnightBlue}
\usepackage{url}

\usepackage[nameinlink]{cleveref}
\creflabelformat{equation}{#1#2#3}
\crefname{equation}{eq.}{eqs.}
\Crefname{equation}{Eq.}{Eqs.}
\usepackage[acronym,smallcaps,nowarn]{glossaries}
\usepackage{listings}
\lstdefinestyle{alp_style}{
    commentstyle=\color{OliveGreen},
    numberstyle=\tiny\color{black!60},
    stringstyle=\color{BrickRed},
    basicstyle=\ttfamily\scriptsize,
    breakatwhitespace=false,
    breaklines=true,
    captionpos=b,
    keepspaces=true,
    numbers=none,
    numbersep=5pt,
    showspaces=false,
    showstringspaces=false,
    showtabs=false,
    tabsize=2
}
\usepackage{amsthm}
\usepackage{multicol}
\usepackage[accepted]{aistats2020}
\usepackage{sectsty}
\usepackage{multirow}
\usepackage{enumerate}
\usepackage[symbol]{footmisc}
\RequirePackage{snapshot}

\newcommand{\bx}{\boldsymbol{x}}
\newcommand{\boldf}{\boldsymbol{f}}
\newcommand{\boldy}{\boldsymbol{y}}
\newcommand{\by}{\boldsymbol{y}}

\newcommand{\bu}{\boldsymbol{u}}

\newcommand{\bomega}{\boldsymbol{\omega}}

\newcommand{\PG}{\mathrm{PG}}
\newcommand{\GP}{\mathrm{GP}}

\newcommand{\KL}{\mathrm{KL}}
\newcommand{\diag}{\mathrm{diag}}

\newcommand{\dd}{\mathrm{d}}
\newcommand{\RR}{\mathbb{R}}

\newcommand{\expec}[2]{\mathbb{E}_{#1}\left[#2\right]}

\newcommand{\ELBO}{\text{ELBO}}
\newcommand{\tr}{\text{tr}}

\newcommand{\half}{\frac{1}{2}}

\newcommand{\laplace}[2]{\mathcal{L}\left\{#1\right\}\left(#2\right)}

\newcommand{\plotdir}[1]{plots/#1}

\newtheorem*{prop*}{Proposition}
\newtheorem{theorem}{Theorem}
\newtheorem*{theorem*}{Theorem}
\newtheorem*{lemma*}{Lemma}
\newtheorem*{corrolary*}{Corrolary}

\newtheorem*{definition*}{Definition}
\sectionfont{\MakeUppercase}
\definecolor{gaussian}{RGB}{0,0,168}
\definecolor{omega}{RGB}{161,21,0}

\newacronym{ACI}{aaci}{\it{automated augmented conjugate inference}}
\newacronym{PDR}{pdr}{positive definite radial}
\newacronym{CDF}{cdf}{cumulative density function}

\hypersetup{citecolor=Violet}
\hypersetup{linkcolor=black}
\hypersetup{urlcolor=MidnightBlue}

\runningtitle{Automated Augmented Conjugate Inference for GP Models}
\date{}
\begin{document}
	\renewcommand*{\thefootnote}{\fnsymbol{footnote}}
	\twocolumn[\aistatstitle{Automated Augmented Conjugate Inference \\for Non-conjugate Gaussian Process Models}
	\aistatsauthor{Th\'eo Galy-Fajou \And Florian Wenzel \And Manfred Opper}
	\aistatsaddress{ Technical University of Berlin \And Google Research\footnotemark \And Technical University of Berlin } ]
%	\twocolumn[\aistatstitle{Automated Augmented Conjugate Inference for Non-conjugate Gaussian Process Models}
%	\aistatsauthor{Th\'eo Galy-Fajou \And Florian Wenzel \And Manfred Opper}
%	\aistatsaddress{ TU Berlin \And TU Kaiserslautern \And TU Berlin } ]
	\begin{abstract}
		We propose \emph{automated augmented conjugate inference}, a new inference method for non-conjugate Gaussian processes (GP) models.
		Our method automatically constructs an auxiliary variable augmentation that renders the GP model conditionally conjugate. Building on the conjugate structure of the augmented model, we develop two inference methods. First, a fast and scalable stochastic variational inference method that uses efficient block coordinate ascent updates, which are computed in closed form. Second, an asymptotically correct Gibbs sampler that is useful for small datasets.
		Our experiments show that our method are up two orders of magnitude faster and more robust than existing state-of-the-art black-box methods.
%		By augmenting the model with auxiliary latent variables we obtain a conditionally conjugate likelihood, leading to closed-form full conditional distributions.
%		We derive two efficient inference m
%
%		Gaussian Processes are a very powerful class of models used in Bayesian inference.
%%		However inferring the exact posterior is only tractable for a Gaussian likelihood.
%		However using a non-conjugate likelihood leads to approximations or sampling methods which can be very slow and unstable.
%		Instead of trying to optimize existing approaches, we propose to simplify the problem by reaching conjugacy via data augmentation. We propose a large class of likelihoods, for which the augmentation can be done automatically, for which we provide both an efficient Gibbs sampling algorithm and a variational inference scheme. Our approach not only outperforms existing methods but has also strong convergence guarantees.
	\end{abstract}
%%%%%%%%%%%%%%%%%%%%%%%%%%%%%%%%%%
\section{Introduction}
\label{sec:introduction}
%%%%%%%%%%%%%%%%%%%%%%%%%%%%%%%%%%

\begin{figure*}[!t]
  \glsresetall
  \centering
  \includegraphics[width=0.9\textwidth]{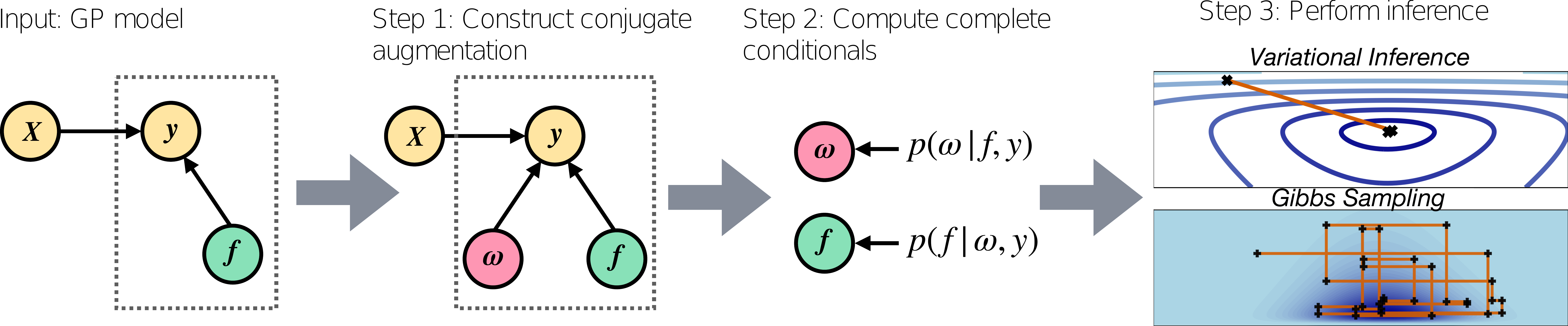}
  \caption{\Gls{ACI} performs automated efficient inference in non-conjugate Gaussian process models. In the first step, \gls{ACI} translates the GP model into an augmented model that is conditionally conjugate. In the second step, the complete conditionals are computed in closed form. In the final step, \gls{ACI} provides two options: (A) fast stochastic variational inference based on coordinate ascent updates, which easily scales to big datasets and (B) an asymptotically exact Gibbs sampler, which provides high quality samples from the true posterior but is limited to smaller datasets.}
  \label{fig:pipe_line}
\end{figure*}

\glsresetall
%* Motivate inference in latent GP models *%
%TODO: That is not a really nice start...
Developing automated yet efficient Bayesian inference methods for Gaussian process (GP) models is a challenging problem that has attracted considerable attention within the probabilisitic machine learning community \citep{salimbeni2018natural, wenzel2019efficient}.
A GP defines a distribution over functions and can be used as a flexible building block to develop expressive probabilistic models.
By choosing an appropriate likelihood function on top of a latent GP, a variety of interesting models is obtained, which are successfully used in several application areas including robotics \citep{GProbot}, facial behavior analysis \citep{eleftheriadis2017tip} and electrical engineering \citep{Pandit_2018}. % and geospatial predictive modeling where they are known as {\em kriging} \citep{stein2012interpolation, JMLR:v19:17-042}.
For instance, using a logistic likelihood leads to a binary GP classification model, and using a Student-t likelihood can be used for robust regression. %cite?

%* What's the problem? *%
The main challenge in these models is to infer the latent GP given a general non-Gaussian likelihood. %TODO I would formulate the other way around
Methods that are more generally applicable often treat the model as a black box and are based on sampling or numerical quadrature, thus, preventing efficient optimization \citep{hensman2015scalable,salimbeni2018natural}.
On the other side. a lot of methods focus on special cases of GP models (i.e. special likelihood functions) by exploiting model specific properties, e.g. binary classification \citep{polson2013bayesian}. %TODO: cite more?

%Or only exploit the model superficially \citep{khan2017conjugate}
 %TODO: Cite more? Khan?

\footnotetext{Work done while at TU Berlin}
\renewcommand*{\thefootnote}{\arabic{footnote}}
\setcounter{footnote}{0}
%* Our goals *%
In this work, we develop \gls{ACI}. \gls{ACI} is an efficient inference framework, which is applicable to a large class of GP models that use a super-Gaussian likelihood\footnote{The definition of the family of super-Gaussian likelihoods is given in Section~\ref{sec:algorithm}.}. It automatically exploits specific properties of the likelihood leading to an inference algorithm that is up to two orders of magnitudes faster than the state of the art.% for all considered likelihood functions.

%* How does it work? Augmentation *%
Our approach builds on an auxiliary variable augmentation of the model: we add a latent variable to the model such that the original model is recovered when this variable is integrated out.
We consider an augmentation that renders the model conditionally conjugate. In a conditionally conjugate model, all complete conditional distributions (the posterior distribution of one random variable given all the others), can be computed in closed form. Moreover, we show that inference in the augmented conditionally conjugate model is much easier than in the original model and demonstrate superior performance over the state of the art.

%* Example Polya-Gamma *%
%For instance, let us consider a GP classification model with a logistic likelihood.
%Without any manual input, \gls{ACI} automatically constructs an appropriate auxiliary variable augmentation. In this case, \gls{ACI} leads to the well-known P\'olya-Gamma augmentation introduced by \citet{polson2013bayesian}.% and and applied in the context of variational inference by \citet{wenzel2019efficient}.
%Of course \gls{ACI} generalizes to a much larger class of models.
%But \gls{ACI} is not limited to this model.
%The method is applicable to many other types of GP models and automatically constructs augmentations, which are similarly efficient as the P\'olya-Gamma approach.

%* Inference Modules + Summary*
Building on the conditionally conjugate augmentation, \gls{ACI} provides two options for inference: a scalable variational inference method based on efficient closed-form coordinate ascent updates and an exact Gibbs sampling method, which is useful on smaller datasets.

Our main contributions are as follows:
\begin{compactitem}
    \item We introduce \gls{ACI}: an automated inference method for GP models with a super-Gaussian likelihood.%, a large family of distributions. <-- sounds weird
%    \gls{ACI} first constructs an auxiliary variable augmentation that renders the model conditionally conjugate and then performs efficient inference in the augmented model.
    \item We propose two inference modules: augmented variational inference, which scales to large datasets containing millions of instances and an exact Gibbs sampler, which is useful for small datasets.
    \item The experiments demonstrate that the augmented variational inference module of \gls{ACI} outperforms the state of the art in terms of speed by up to two orders of magnitude while being competitive in terms of prediction performance.
    The Gibbs sampler module leads to a much better efficient sample size while still being up to ten times faster than Hamiltonian Monte Carlo.
\end{compactitem}
%%%%%%%%%%%%%%%%%%%%%%%%%%%
The paper is structured as follows: Section~\ref{sec:method} gives a high-level overview about our novel inference method \gls{ACI}.
In Section~\ref{sec:algorithm}, we provide a detailed discussion of the algorithm and proof that our approach indeed leads to conditionally conjugate models.
%Section~\ref{sec:theory} \flo{fill in}.
We discuss related work in Section~\ref{sec:related_work} and show our experimental results in Section~\ref{sec:experiments}.
Finally, Section~\ref{sec:conclusion} concludes and lays out future research directions. Our source code for the experiments is included in a gitgub repository\footnote{\url{https://github.com/theogf/AutoConjGP_Exp}}.

%\input{introduction_v1}
%%%%%%%%%%%%%%%%%%%%%%%%%%%%%%%%%%
\vspace{-0.3cm}
\section{Automated Augmented Conjugate Inference}
\label{sec:method}
\glsresetall
%%%%%%%%%%%%%%%%%%%%%%%%%%%%%%%%%%

%* Introduce GP Model *%
Let $X=(\bx_1, \dots, \bx_n)^\top \in \RR^{n\times d}$ be a matrix of data points and $\by=(y_1, \dots, y_n) \in \RR^n$ the corresponding target values. The goal is to learn a mapping from the input points to the target values via a latent function $f$. We assume a prior GP distribution (with mean prior $\boldsymbol{\mu}_0$ and covariance function $k(x,x')$) on the latent function and the data labels
 $\by=(y_1, \dots, y_n)$ are connected to $f$ via a factorizable likelihood
 \begin{align*}
   p(f) = \GP(f | \boldsymbol{\mu}_0,k),\quad   p(\boldy | f, X) = \prod_{i=1}^n p(y_i | f(\bx_i)).
 \end{align*}
 %TODO add again later?
% We consider GP models where the likelihood function is in the family of \emph{super-Gaussian likelihoods}\footnote{The definition of the family of super-Gaussian likelihoods is given in Section~\ref{sec:algorithm}.}.
% A large class of known problems can be well modeled with the above assumptions including binary GP classification using a logistic likelihood\citep{polson2013bayesian}, the Bayesian support vector machine (SVM) \citep{henao2014bayesian,wenzel2017bayesian}, robust GP regression using a Student-t likelihood \citep{jylanki2011robust} or a Laplace likelihood.
% But it is also possible to build entirely novel likelihoods as e.g. the Matern likelihood (see Section~\ref{sec:experiments}).
%
% Explain the problem with Inference
The key inference challenge in the GP models is to compute the posterior distribution of the latent function
\begin{align*}
    p(f|y) = \frac{p(y|f)p(f)}{\int p(y|f)p(f) \dd y}.
\end{align*}
This is a challenging problem.
Inference in GP models scale cubically in the number of data points and is intractable for non-Gaussian likelihoods.

%* Existing Metdhods <-> Intro *&
%Scalable approximate inference methods exist \citep[e.g.][]{hernandez2016scalable,salimbeni2018natural}, but they typically build on approximating the variational objective by sampling or numerical quadrature, thus, preventing efficient and stable optimization.
%Recently, more efficient inference methods have been proposed for a few particular GP models that do not rely on numerical approximations.
%For  instance, \citet{wenzel2019efficient} propose an augmented variational inference method for GP classification and ... \flo{Cite work from other authors here?}
%These approaches, however, are tailored to one particular likelihood.

%* Our goal <-> Intro *%
Ideally, we would like an efficient inference method that is not hand-tailored to a specific type of likelihood and hence allows for experimenting with different types of GP models on big datasets in a scalable manner.
Thus, we need a flexible inference method that works for a large class of likelihoods, is fast and ideally does not involve inefficient black box approaches as approximating the objective by sampling.
\subsection{Automated Augmented Conjugate Inference}
We introduce the \gls{ACI} to achieve this goal. \gls{ACI} accelerates
training of GP models whose likelihood is in the family of super-Gaussian likelihood functions.

%* How it works <-> Intro *%
% TODO Cite:
\gls{ACI} translates the intractable non-conjugate model into an easier, conditionally conjugate model by adding auxiliary random variables to the model.
%An auxiliary variable is an additional latent variable that is added to a model such that the original model is recovered when this variable is marginalized out.
Inference in conditionally conjugate models is a classic and well-studied problem \citep{bishop2006pattern}.
Because of the special structure of conditionally conjugate models, many efficient inference methods exist \citep{wang2013variational}. %TODO eventually remove
Based on the automatically constructed augmentation, we propose an efficient variational inference method using coordinate ascent updates and a Gibbs sampler.

%The augmented model has the appealing property that inference of the latent function $f$ reduces to a much easier GP regression problem with a Gaussian likelihood.
%In this case, the posterior of $f$ given the auxiliary variables can be computed in closed form.

\parhead{The inference pipeline of \gls{ACI}.}
\gls{ACI} consists of three steps. In the first step, a conjugate augmentation of the model is constructed by adding auxiliary variables $\omega$ to the model. Then, the complete conditional distributions of the latent function $f$ and auxiliary variables $\omega$ are computed. 
In the final step, we provide two options to perform inference.

%Option (A): \emph{Augmented variational inference} for scalable \emph{approximate} inference on big datasets and  \emph{Gibbs sampling} which carries out \emph{exact} inference but is limited to smaller datasets.

The \emph{variational inference} (VI) module of \gls{ACI} performs block coordinate ascent updates, computed in closed form.
The updates are much more efficient than ordinary Euclidean gradient updates, which are used in most previous approaches. % and our algorithm is up to two orders of magnitude faster than the state of the art. 
The \emph{Gibbs sampling} module of \gls{ACI} builds on the complete conditional distributions and provides exact samples from the true posterior.
For each type of likelihood, the sampler is automatically constructed.

The inference pipeline of \gls{ACI} is summarized in Fig.~\ref{fig:pipe_line}. In the following, we give an overview of how each module of our inference pipeline works and provide the details in Section~\ref{sec:algorithm}.

% %* Inference Pipeline *%
% \begin{tabular}{l l}
%   %\hline
%   %\multicolumn{2}{l}{Augmented Conjugate Inference} \\
%   Input:  & \textbf{GP model}\\
%           & \quad $p(y|f) p(f)$\\
%   Step 1: & \textbf{Construct conjugate augmentation}\\
%           & \quad $p(y|,\omega, f) p(f) p(\omega)$\\
%           & \qquad \it{The auxiliary variable augmentation is}\\
%           & \qquad \it{automatically constructed, see Sec.~\ref{subsec:augmentation}.}\\
%   Step 2: & \textbf{Compute complete conditionals}\\
%           & \quad $p(f|\omega, y)$, \quad $p(\omega|f, y)$\\
%           & \qquad \it{Computed in closed form, see Sec.~\ref{subsec:complete_conditionals}}\\
%   Step 3: & \textbf{Perform inference}\\
%           & \quad Method (A): Augmented SVI\\
%           & \qquad \it{Using efficient CAVI updates, see Alg.~\ref{alg:VI}.}\\
%           & \quad Method (B): Gibbs Sampling\\
%           & \qquad \it{Sampling scheme shown in Alg.~\ref{alg:Gibbs}.}\\
% \end{tabular}

\parhead{(1) Augmenting the model.}
The first step of our inference framework constructs an \emph{auxiliary variable augmentation} that renders the model \emph{conditionally conjugate}. Our augmentation approach finds a Gaussian scale mixture representation of the intractable likelihood
% \begin{align}
% p(y|f) = \int \underbrace{p(y|f,\omega)}_{\text{Gaussian in $f$}} \underbrace{p(\omega)}_{\text{Augment.}} \dd\omega, \label{eq:scalemixture}
% \end{align}
\begin{align}
p(y_i|f_i) = \int p(y_i|f_i,\omega_i) p(\omega_i) \dd\omega, \label{eq:scalemixture}
\end{align}
where $p(y_i|f_i,\omega_i)$ is an unnormalized Gaussian distribution in $f_i$ with precision $\omega_i$
and $p(\omega_i)$ is the prior distribution of the auxiliary variable. The construction of the distribution $p(\omega)$ is based on an inverse Laplace transformation and is discussed in Section~\ref{subsec:augmentation}.

Building on Eq.~\ref{eq:scalemixture}, we augment the GP model by a set of auxiliary variables  $\bomega=(\omega_1, \dots, \omega_n)$ leading to the augmented joint distribution
\begin{align}
        p(\by, \boldf, \bomega) = \prod_i p(y_i|f_i, \omega_i) p(\omega_i) p(\boldf), \label{eq:augmented_model}.
\end{align}
The auxiliary variable augmentation is constructed in a way such that the augmented model is \emph{conditionally conjugate}, i.e. the complete conditional distributions $p(\bomega|\boldf, \by)$ and $p(\boldf| \bomega, \by)$ are in the same family as their associated priors.

% Give short reference that in the case of GP classification with a logistic likelihood our procedure would \emph{automatically} construct a P\'olya-Gamma augmentation that was \emph{manually} derived for that special case in previous work \citep{polson2013bayesian} and applied in setting of variational inference by \cite{wenzel2019efficient}. \flo{improve! more in sec 3}

% \begin{enumerate}[(1)]
%     \item The original model is restored by marginalizing $\bomega$, i.e. $p(\by,\boldf) = \int p(\by, \boldf, \bomega) \dd \bomega$.
%     \item The augmented model is \emph{conditionally conjugate} and the complete conditional distributions are computed in closed form.
%     \item We can perform efficient inference in the augmented model.
% \end{enumerate}
% In a conditionally conjugate model, the complete conditional distributions are in the same family as the prior distributions...
% Can be computed efficiently and we are the basis for inference modules...

%* Explain why this a advantageous *%

\parhead{(2) Computing the complete conditionals.}
The complete conditionals of $\boldf$ and the auxiliary variables $\omega_i$ are computed in closed form and are given by
\begin{align*}
\begin{aligned}
  p(\boldf|\by,\bomega) =& \mathcal{N}\left(\boldf| \boldsymbol{\mu} ,\boldsymbol{\Sigma}\right)\\
  p(\omega_i|f_i,y_i) =& \pi_{\varphi}\left(\omega_i| c_i\right),
\end{aligned}
\end{align*}
where $\varphi$ is a function determined by the type of the likelihood (see Eq.~\ref{eq:likelihood}) and the parameters $\boldsymbol{\mu}, \boldsymbol{\Sigma}, c_i$ have closed-form expressions and are described in Section~\ref{subsec:complete_conditionals}.
The distribution family $\pi_{\varphi}\left(\omega| c\right)$ is derived by an exponential tilting %\com{TMI at this stage}
of the prior distribution $p(\omega)$ and is discussed in Section~\ref{subsec:complete_conditionals}.

\parhead{(3a) Augmented variational inference.}
In step 3, \gls{ACI} provides two options to perform inference. We first discuss the variational inference module, which approximates the posterior by a variational distribution and easily scales to big datasets.

We assume a mean-field variational distribution, where the latent GP $\boldf$ and the auxiliary variables $\bomega$ are decoupled, i.e. $q(\boldf,\bomega) = q(\boldf)q(\bomega)$.
The optimal variational distribution of $\bomega$ naturally factorizes, i.e. $q(\bomega) = \prod_i q(\omega_i)$. Following standard results \citep{bishop2006pattern} the variational distributions can be iteratively optimized by the block-coordinate ascent updates:
\begin{align}
  \begin{split}
    q(\boldf)&\propto \exp \left(\expec{q(\bomega)}{\log p(\boldf|\bomega,\by)}\right)\\
    q(\omega_i)&\propto \exp \left( \expec{q(\boldf)}{\log p(\omega_i|\boldf,\by)}\right).
  \end{split}
  \label{eq:cavi_updates}
\end{align}
In Section~\ref{subsec:inference}, we show that these updates are given in closed form and can be computed efficiently without resorting to numerical methods.
%
%* Comment on SVI *%
To scale to big datasets we employ SVI \citep{hoffman2013stochastic} and replace the original latent GP $f$ by \citet{titsias2009variational} sparse approximation building on inducing points .
% \cite{JMLR:v14:hoffman13a} have shown in the setting, where the complete conditionals are in the exponential family, that the coordinate updates~\eqref{CHAP2_eq:cavi_updates} can be directly interpreted as natural gradient updates with a learning rate of one.
% In SVI, the variational objective is optimized via stochastic optimization based on stochastic natural gradients.
% In each iteration, we use mini-batches of the data and obtain a noisy version of the natural gradient. In this setting, learning rates slightly less than one have to be chosen.

%* Why are the updates more efficient *%

%* On the quality of the augmentation? *%

\parhead{(3b) Exact inference via Gibbs sampling.}
Building on the conditionally conjugate augmentation, it is straightforward to derive a Gibbs sampler. In order to sample from the exact posterior, we alternate between drawing a sample from each complete conditional distribution
\begin{align*}
	\bomega^t &\sim p(\bomega|\boldf^{t-1}, \boldy),\\
    \boldf^t &\sim p(\boldf|\bomega^{t}, \boldy).
\end{align*}
The augmented variables are naturally marginalized out and the latent GP samples $\{\boldf^t\}$ will be from the true posterior $p(\by|\boldf)$.
As we empirically show in Section~\ref{sec:expgibbs}, the Gibbs sampler leads to very fast mixing and outperforms standard Hamiltonian Monte Carlo sampling.

%Since the complete conditional distribution of the latent GP $\boldf$ \com{this sentence does not make sense, what did you mean} we show how \gls{ACI} automatically constructs an efficient sampler for the complete conditional distribution $p(\bomega|\boldf_{t}, \boldy)$. The sampling scheme is based on the idea \flo{insert}.

%%%%%%%%%%%%%%%%%%%%%%%%%%%%%%%%%%
\section{Algorithm Details} %Description?
\label{sec:algorithm}
%%%%%%%%%%%%%%%%%%%%%%%%%%%%%%%%%%
\glsresetall

Here we provide the details on the \gls{ACI} algorithm.
We start by specifying the class of GP models that we consider in our framework. We then discuss the technical details of \gls{ACI} and proof that the automatically constructed augmentation indeed leads to a conditionally conjugate model.

\parhead{GP Models with a super-Gaussian likelihood.}
  \gls{ACI} can be applied to GP models, where the likelihood is within the class of super-Gaussian likelihoods. A super-Gaussian likelihood is of the form
	\begin{align}
	p(\by|\boldf;\theta) =& C(\theta)e^{g(\by;\theta)^\top \boldf}\varphi(||h(\boldf,\by)||_2^2), \label{eq:likelihood}
	\end{align}
	where $\theta$ are hyperparameters of the likelihood, $C(\theta)$ is the normalizing constant, $g(\by;\theta)$ is an arbitrary function, $\varphi$ is a \emph{\gls{PDR} function}\footnote{$\varphi$ is a positive definite radial function if $\varphi(r)$ is completely monotone for all $r\ge 0$ and $\lim_{r\rightarrow 0} \varphi(r)= 1$.},
  %$\|\cdot\|_2$ is the $l^2$-norm
	and $h$ is a linear function in $\boldf$, such that we can write
	\begin{align}
	||h(\boldf,\by)||_2^2 =& \alpha(\by,\theta) - \beta(\by,\theta)^\top \boldf + \gamma(\by,\theta)||\boldf||^2_2, \label{eq:h}
	\end{align}
	where $\alpha,\beta,\gamma$ are arbitrary functions.
	We omit $\theta$ in the later derivations for clarity.

  Many interesting models
  %---including GP classification, the Bayesian SVM \cite{henao2014bayesian, wenzel2017bayesian} and robust GP regression---
  are instances of super-Gaussian likelihood GP models.
  In Table~\ref{tab:likelihoods}, we present several likelihood functions with their corresponding parameter settings of the super-Gaussian likelihood as given in Eq.~\ref{eq:likelihood}.

  \begin{table*}
	\begin{center}
		\begin{tabular}{c|c|c|c|c}
			Likelihood & Full form & $g(f,y)$ & $h(f,y)$ & $\varphi(r)$\\\hline
       %TODO: Deleted dependency of \varphi(r;\theta) to avoid confusion, since the actual forms don't involve \theta
			Student-t &$\frac{\Gamma\left(\frac{\nu+1}{2}\right)}{\sqrt{\nu\pi}\sigma\Gamma\left(\frac{\nu}{2}\right)} \left(1+\frac{(y-f)^2}{\nu\sigma^2}\right)^{-\frac{\nu+1}{2}}$  & 0 & $\frac{f-y}{\sigma}$ & $\left(1+\frac{r}{\nu}\right)^{-\frac{\nu+1}{2}}$\\
			%		Multivariate Student-T & $\frac{\Gamma[(\nu+p)/2]}{\Gamma(\nu/2)\nu^{p/2}\pi^{p/2}\sqrt{|\Sigma|}}\left[1+\frac{1}{\nu}(f-y)^\top\Sigma^{-1}(f-y)\right]^{-(\nu+p)/2}$ & 0 & $(f-y)^\top\Sigma^{-1}(f-y)$& $(1+\frac{r}{\nu})^{-(\nu+p)/2}$\\
			Laplace & $\frac{1}{2\beta}\exp\left(-\frac{|y-f|}{\beta}\right)$ & $0$ & $f-y$ & $\exp\left(-\frac{\sqrt{r}}{\beta}\right)$\\
%			Generalized Cauchy & $\frac{\alpha\Gamma(\nu+1/\alpha)}{2\Gamma(1/\alpha)\Gamma(\nu)}(1+|\frac{y-f}{\sigma}|^\alpha)^{-\nu-1/\alpha}$ & 0 & $f-y$ & $\left(1+\frac{r^{\frac{\alpha}{2}}}{\sigma^\alpha}\right)^{-\nu-1/\alpha}$\\
			Logistic & $\frac{1}{2}\exp\left(\frac{yf}{2}\right)\cosh^{-1}\left(\frac{|yf|}{2}\right)$ & $\frac{yf}{2}$ & $\frac{f}{2}$ &  $\cosh^{-1}\left(\sqrt{r}\right)$\\
			Bayesian SVM & $\exp\left((yf-1)-|1-yf|\right)$& $yf$ & $1-yf$ & $\exp(-\sqrt{r})$\\
			Matern 3/2 & $\frac{\sqrt{3}}{4\rho}(1+\frac{\sqrt{3}|y-f|}{\rho})\exp(-\frac{\sqrt{3}|y-f|}{\rho})$ & 0 & $f-y$ & $(1+\frac{\sqrt{3r}}{\rho})\exp(-
			\frac{\sqrt{3r}}{\rho})$\\
		\end{tabular}
	\end{center}
	\caption{Many interesting GP models are members of the super-Gaussian likelihood family introduced in Section~\ref{sec:algorithm}. We display the full likelihood and the corresponding terms of the super-Gaussian likelihood as described in Eq.~\ref{eq:likelihood}. Some models were already considered independently but our approach provides a unified view.}
	\label{tab:likelihoods}
\end{table*}

  \parhead{Constructing new likelihoods.} Using Eq.~\ref{eq:likelihood}, we can also construct novel likelihood functions based on existing kernel functions. In this paper we propose the Matern 3/2 likelihood.

 %  In the future, the dependence in $\theta$ is omitted for clarity.
 % PDR functions have a lot of properties, described in \citet{merkle2014completely}, that can be used for considering a much richer class of distributions.\\Given $\{\varphi_i\}$ a collection of PDR functions:%
 % \begin{enumerate}
 % 	%		\item If $\varphi_i:\mathbb{R}^d \in\text{PDR}$, then $\varphi_i:\mathbb{R}^k \in \text{PDR}$, $\forall k < d$
 % 	\item For $\alpha_i \geq 0$, $\sum_{i=1}^n \alpha_i \varphi_i(r)$ is PDR.
 % 	\item $\prod_i \varphi_i(r)$ is PDR.
 % 	\item For $\lambda>0$, $\exp\left(\lambda \varphi(r)\right)$ is PDR.
 % 	\item For $\lambda>0$, $\varphi^\lambda(r)$ is PDR.
 % %	\item A convolution of PDR functions is a PDR function.
 % \end{enumerate}
 % Using properties 1 and 2 allows for a much richer class of distributions by combining existing PDR functions.
 % Property 2 and 4 can also be generalized to distributions of the form (\ref{eq:general_likelihood}), as the operations would conserve the form of (\ref{eq:general_likelihood}).
 %
 %
 % Given these properties and the equation (\ref{eq:augmentation}), one can understand that the Gaussian is a basis function for all PDR functions. All other PDR functions are a linear combination of the radial basis function (RBF).

\subsection{Step 1: Conjugate augmentation}
\label{subsec:augmentation}
Given the likelihood of the model, \gls{ACI} constructs a conditionally conjugate auxiliary variable augmentation as follows.
We first define a family of distribution $\pi_\varphi(\omega|c)$, which will be useful for constructing the augmentation.
%covers the prior distribution of the auxiliary variables $p(\omega)$ as well as the complete conditional distribution $p(\omega|f,y)$.

For the case $c=0$, the distribution $\pi_\varphi(\omega|0)$ is defined by the inverse Laplace transform of $\varphi(\cdot)$,
\begin{align}
\pi_\varphi(\omega|0) = \mathcal{L}^{-1}\left\{\varphi(\cdot)\right\}(\omega).\label{eq:invlapomega}
\end{align}
%which only depends on the \gls{PDR} function $\varphi(\cdot)$ of the super-Gaussian likelihood (see Eq.~\ref{eq:likelihood}).

%It turns out that we do not need to compute the inverse Laplace transformation explicitly. In the variational inference algorithm (Section~\ref{subsec:VI}) we only have to compute the first moment of  $p(\omega)$, which boils down to computing first derivative of $\varphi(\cdot)$. This can be cheaply done via automatic differentiation. %TODO: Repeat that in the VI section?
%\com{I would leave this out for the VI part}
%For the Gibbs sampling method (Section~\ref{subsec:Gibbs}), we propose a sampler for $\pi_\varphi(\omega|c)$ that builds on evaluating the inverse Laplace transformation point-wise. That can be done efficiently by... \flo{check!}

The inverse Laplace is the inverse mapping of the Laplace transformation and can be computed by the Bromwich integral formula\footnote{The inverse Laplace transformation of a function $\varphi(\cdot)$ can be computed by $\mathcal{L}^{-1}\left\{\varphi(\cdot)\right\}(\omega) = \lim_{T\rightarrow\infty} \frac{1}{2\pi i}\int_{b-iT}^{b+iT} e^{r\omega}\varphi(r)dr$, where $b$ can be arbitrarily chosen but has to be larger than the real part of all singularities of $\varphi$.} \citep{debnath2014integral} and it defines a valid density in our setting (see proof of Theorem~\ref{thm:augmentation}).
Remarkably, we will see that for the final updates of our algorithm, we do not need to compute the inverse Laplace transformation explicitly.

We generalize the base distribution $\pi_\varphi(\omega|0)$ by applying an exponential tilting:
\begin{align}
\pi_\varphi(\omega|c) = \frac{e^{-c^2 \omega}\pi_\varphi(\omega|0)}{\varphi(c^2)}, \label{eq:piomega}
\end{align}
where $c\in \RR$.
%Note that the Laplace transform of $\pi_\varphi(\omega|c)$ corresponds to an exponential tilting : $\laplace{\pi_\varphi(c)}{r} = \varphi(r+c^2)/\varphi(c^2)$

\begin{theorem} A GP model with a super-Gaussian likelihood (of the form of Eq.~\ref{eq:likelihood}) is rendered \textbf{conditionally conjugate} by the auxiliary variable augmentation ${p(\by,\boldf,\bomega;\theta) = p(\by|\boldf,\bomega;\theta) p(\boldf) p(\bomega)}$. The augmented likelihood is
\begin{align*}
	p(\by|\boldf,\bomega;\theta) &= C(\theta)\exp\left(g(\by;\theta)^\top \boldf - ||h(\boldf,\by)||_2^2\bomega\right)
\end{align*}
and the prior distribution of the auxiliary variables is
\begin{align*}
  p(\bomega) =\pi_{\varphi}\left(\bomega| 0\right).
\end{align*}
\label{thm:augmentation}
\end{theorem}
\parhead{Proof:}
%\flo{This makes no sence since pdr is only defined for real valued functions so far.}\\
We first apply Schoenberg's theorem \citep{schoenberg1938metric}, which states that a function $\mathbb{R}^d\ni \boldsymbol{x}\rightarrow \varphi(\|\boldsymbol{x}\|_2^2)$ is a \gls{PDR} function for any dimension $d>0$ if and only if $\varphi(r)$ is a completely monotone function on the domain $r\geq0$.

%\flo{Take this?}\\
%We first apply Schoenberg's theorem \citep{schoenberg1938metric}, which states that the \gls{PDR} function $\varphi(\cdot)$ of the super-Gaussian likelihood is a  completely monotone function on the domain $r\geq0$.

A completely monotone function $\varphi(\cdot)$ has the property that it is infinitely differentiable and its derivatives have an alternating sign \citep{bernstein1929fonctions}, i.e.
\begin{align}
	(-1)^k\varphi^{(k)}(r) > 0,\quad r\in[0,+\infty),\; k=0,1,2,\ldots.
\end{align}
As a direct consequence, $\varphi(\cdot)$ is a positive, decreasing, and convex function and the first derivative of $\varphi(\cdot)$ is a concave function.

% For definition \citet{bernstein1929fonctions}, a function $\varphi(.)$ is completely monotone function on $[0,\infty)$, if it is $C^\infty$ and satisfies :
% \begin{align}
% 	(-1)^k\varphi^{(k)}(t) > 0,\quad t\in[0,+\infty), k=0,1,2,\ldots.
% \end{align}
% One direct consequence is that $\varphi(\cdot)$ is positive, decreasing and convex, with a concave first derivative.

Building on these properties, \citet{widder1946laplace}~states that we can rewrite $\varphi(\|h(f,y)\|_2^2)$ as a Gaussian scale-mixture
\begin{align}
\varphi\left(\|h(f,y)\|_2^2\right)= \int_0^\infty e^{-\|h(f,y)\|_2^2 \omega} d\mu(\omega),\label{eq:scale-mixture}
\end{align}
with respect to a Borel measure $\mu(\omega)$.
We apply the monotone convergence theorem \citep{yeh2006real}, which gives that $\mu(\omega)$ is even a probability measure iff $\lim_{r\rightarrow 0}\varphi(r)=1$. Since we have a probability measure, we write $d\mu(\omega) = p(\omega)d\omega$ and which leads to the equality $\varphi(r)=\laplace{p(\omega)}{r}$, where $\mathcal{L}$ denotes the Laplace transformation. The inverse Laplace transformation gives the density of the auxiliary variable ${p(\omega)=\mathcal{L}^{-1}\left\{\varphi(r)\right\}(\omega)=\pi_\varphi(\omega|0)}$.

Therefore we can rewrite the super-Gaussian likelihood Eq. \ref{eq:likelihood} as :
\begin{align}
p(y|f)=C(\theta)\int_0^\infty e^{-g(y)f-\|h(f,y)\|_2^2\omega}p(\omega)d\omega.
\end{align}
Adding the auxiliary variable $\omega$ with prior $p(\omega)$ to the model, we obtain the augmented likelihood	${p(\by|\boldf,\bomega;\theta) = C(\theta)\exp\left(g(\by;\theta)^\top \boldf - ||h(\boldf,\by)||_2^2\bomega\right)}$.

Since the function $g(\by;\theta)^\top \boldf - ||h(\boldf,\by)||_2^2\bomega$ is by definition quadratic in $\boldf$ the augmented likelihood is proportional to an (unnormalized) Gaussian distribution in $\boldf$, hence, conditionally conjugate in $\boldf$.\\
For the augmented variable $\omega_i$, the likelihood $p(y|\omega,f)$ act as an exponential tilting of $p(\omega)$ and the full conditional in $\omega$ will stay in the same family of distributions.
QED.

%\parhead{Remark.}
%\flo{Maybe remove here}
%% TODO: short reference to intro
%% TODO: in related work say we include bayes svm and x-gpc as special cases
%In the case of GP classification with a logistic likelihood our procedure would \emph{automatically} construct a P\'olya-Gamma augmentation that was \emph{manually} derived for that special case in previous work \citep{polson2013bayesian} and applied in setting of variational inference by \cite{wenzel2019efficient}.

\subsection{Step 2: Complete Conditionals}
\label{subsec:complete_conditionals}
 Since the augmented model (Section~\ref{subsec:augmentation}) is conditionally conjugate, the complete conditional distribution are in the same family as their associated prior distributions and are given in closed form.

\begin{theorem}
  The \textbf{complete conditional distributions} of the augmented model presented in Section~\ref{subsec:augmentation} are given by
  \begin{align}
  \begin{aligned}
    p(\omega_i|f_i,y_i) =& \pi_{\varphi}\left(\omega_i| \|h(f_i,y_i)\|_2\right),\\
    p(\boldf|\by,\bomega) =& \mathcal{N}\left(\boldf| \boldsymbol{\mu} ,\boldsymbol{\Sigma}\right),
  \end{aligned}\label{eq:fullconditionals}
  \end{align}
  where $\boldsymbol{\Sigma} = \left(\diag\left(2\bomega\circ \gamma(\by)\right) + K^{-1}\right)^{-1}$ and $\boldsymbol{\mu} = \boldsymbol{\Sigma}\left(g(\by)+\bomega\circ \beta(\by) + K^{-1}\boldsymbol{\mu}_0\right)$, $\circ$ denotes the Hadamard product and the function $h(\cdot)$ is given by the form of likelihood (see  Eq.\ref{eq:h}).
  \label{thm:fullcond}
  %, $\gamma$ and $\beta$ are applied element-wise.
\end{theorem}\vspace{-0.3cm}
%
%Note that given this definition, it immediately follows that $p(\by|\boldf,\bomega)p(\bomega) \propto \pi_\varphi(\bomega|\|h(\boldf,\by)\|_2)$ and that $\mathcal{L}\left[\pi_\varphi(c)\right](r) = \varphi(c^2+r)/\varphi(c^2)$
The proof is given in Appendix~\ref{appendix:fullcond}

\subsection{Step 3: Efficient inference}
\label{subsec:inference}
In the final step of our inference pipeline, we leverage the conditionally conjugate structure of the augmented model and derive two inference methods. First, we propose a scalable stochastic variational inference (SVI) method that builds on efficient block coordinate ascent updates (CAVI) updates,  computed in closed form. Second, we develop a Gibbs sampling scheme that generates samples from the exact posterior.

\subsubsection{Augmented variational inference}
%A common method for scaling a model to large datasets or simply get a faster convergence is to use variational inference (VI) techniques \citep{blei2017variational}.
%VI aims at finding a distribution (from a given family) approximating the posterior distribution by optimizing a distance.
\label{subsec:VI}
%Building on the conditionally conjugate representation of our model, deriving an efficient variational inference scheme is straightforward.
We implement the classic stochastic variational inference (SVI) algorithm for conditionally conjugate models described by \citet{hoffman2013stochastic}, which builds on block coordinate ascent updates. The updates can be interpreted as natural gradient updates and are much more efficient than ordinary Euclidean gradient updates \citep{amari1998natural}.

\parhead{Variational approximation.}
We approximate the posterior distribution of the latent GP values by
assuming a decoupling between $\boldf$ and $\bomega$.
%$p(u,\bomega|\by) \approx q(\bu,\bomega) = q(\bu, \bomega)$.
%q(\bu)q(\blambda)q(\bn)q(\bomega)q(\tilde \bomega).
The family of the optimal variational distribution can be easily determined
by averaging the complete conditionals in log-space, as given in Eq.~\ref{eq:cavi_updates}~\citep[see e.g.][]{blei2017variational}.
From the above decoupling assumption, it follows that the optimal variational posterior is in the variational family
\begin{align}
q(\boldf,\bomega) = q(\boldf)\prod_{i=1}^Nq(\omega_i), \label{eq:meanfield}
\end{align}
where $q(\boldf)=\mathcal{N}\left(\boldf| \boldsymbol{m},\boldsymbol{S
}\right)$ and $q(\omega_i)=\pi_{\varphi}(\omega_i| c_i)$ and $\boldsymbol{m},\boldsymbol{S}$ and $\boldsymbol{c}$ are the \emph{variational parameters}.

\parhead{Variational updates.}
%Remarkably, the coordinate ascent updates, which are given in Eq.~\ref{eq:cavi_updates} can be computed in closed form without needing to know $q(\omega)$ in closed-form, i.e. evaluating the inverse Laplace transformation of $\varphi$ (see Eq.~\ref{eq:invlapomega}).
%
We start with deriving the variational updates for the variational Gaussian distribution,
\begin{align*}
	q(\boldf) &\propto \exp\left[\expec{q(\bomega)}{\log p(\boldf|\bomega,\by)}\right]\\
	&\propto\exp\left[\sum_i g(y_i) f_i - \|h(f_i,y_i)\|_2^2\expec{q(\omega_i)}{\omega_i}\right]p(\boldf)
%	&\propto\exp\left[\sum_i g(y_i) f_i - \|h(f_i,y_i)\|_2^2\overline{\omega}_i\right]p(\boldf)
\end{align*}
Computing the variational updates of $q(\boldf)$ boils down to computing the first moment of $\bomega$.
Remarkably, the moments %and cumulants % Only to refer to moments here, and introduce cumulants in the theory section if needed
of $\pi_\varphi$ can be computed without computing the closed-form density of $\pi_\varphi$ explicitly,  i.e. without evaluating the inverse Laplace transformation of $\varphi$ (Eq.~\ref{eq:invlapomega}).

The moments can be computed by differentiating the moment generating function, which is itself a Laplace transform.
For our algorithm, we only need the first moment of $\omega$, which is given by
\begin{align*}
\expec{q(\omega)}{\omega} =& \left.\frac{d\laplace{q(\omega)}{-t}}{dt}\right|_{t=0}= -\frac{\varphi'(c^2)}{\varphi(c^2)}=\overline{\omega},
%\frac{d\log(\varphi(c^2))}{dc} = m
\end{align*}
which can be cheaply computed via automatic differentiation.

%For higher moments (cumulants) the detailed derivations can be found in Appendix~\ref{appendix:moment}. %TODO: Move this to theory section

%The updates for the parameters of $\boldf$ are straightforward to obtain since it has now a conjugate likelihood and are described in the algorithm \ref{alg:VI}

The updates for the variational distribution of the auxiliary variables $q(\bomega)$ are computed as follows.
% \begin{align*}
% 	q(\bomega) &\propto \exp\left[\expec{q(\boldf)}{\log p(\bomega|\boldf,\by)}\right]\\
% 	&\propto \exp\left[ -\expec{q(\boldf)}{h(\boldf,\by)^2}\bomega+\log p(\bomega)\right]\\
% 	&=\frac{e^{-\boldc^2 \bomega}p(\bomega)}{\varphi(\boldc^2)}.\label{eq:vardist}
% \end{align*}
\begin{align*}
	q(\omega_i) &%\propto \exp\left[\expec{q(f_i)}{\log p(\omega_i|f_i,y_i)}\right]\\
	\propto \exp\left[ -\expec{q(f_i)}{\|h(f_i,y_i)\|_2^2}\omega_i+\log p(\omega_i)\right]\\
	&\propto \exp\left(-\expec{q(f_i)}{\|h(f_i,y_i)\|_2^2} \omega_i\right)p(\omega_i)\\
	&= \pi_\varphi(\omega_i|\sqrt{\expec{q(f_i)}{h(f_i,y_i)^2}}).\label{eq:vardist}
\end{align*}
We get then the update  ${c_i=\sqrt{\expec{q(f_i)}{\|h(f_i,y_i)\|_2^2}}}$, which can be easily computed in closed form since $\|h(f_i,y_i)\|_2^2$ is a quadratic function of $f_i$.

The coordinate ascent variational inference (CAVI) method is summarized in Algorithm~\ref{alg:VI}.

%		\begin{align}
%			\Sigma =& \left(2\diag(\vec{m})+K^{-1}\right)^{-1}\\
%			\mu =& \Sigma\left(K^{-1}\mu_0+g(y,\theta)+\vec{m}\circ \beta(y,\theta)\right)
%		\end{align}

\newlength{\textfloatsepsave}
\setlength{\textfloatsepsave}{\textfloatsep}
\setlength{\textfloatsep}{0.10in}
\begin{algorithm}
  \caption{Augmented Variational Inference}
  \label{alg:VI}
  \begin{algorithmic}
    %	\KwData{$\mathbf{X},\by$}
    	\STATE {\bfseries Input:} Data $(\mathbf{X},\by)$, GP model $p(\by|\boldf)$, kernel $k$ %\boldsymbol{\mu}_0,
    	\STATE {\bfseries Output:} Approximate posterior {$q(f) =
      \mathcal{N}(f\mid\boldsymbol{m},\boldsymbol{S})$}
    	\FOR{iteration $t=1,2,\dots$,}
        \STATE \verb|# Local updates:|
        \FOR{$i \in 1:N$}
    		  \STATE $c_i=\sqrt{\expec{q(f)}{h(f_i,y_i)^2}}$
    		    \STATE $\overline{\omega}_i = \expec{q(\omega_i)}{\omega_i} =  -\varphi'(c_i^2)/\varphi(c_i^2)$
        \ENDFOR
        \STATE \verb|# Coordinate ascent updates (CAVI):|
    		\STATE $\boldsymbol{S} \leftarrow \left(\diag\left(2\overline{\bomega} \circ \gamma(\by)\right) + K^{-1} \right)^{-1}$
    		\STATE $\boldsymbol{m} \leftarrow \boldsymbol{S}\left(K^{-1}\mu_0 + g(\by) + \overline{\bomega} \circ \beta(\by)\right)$
    \ENDFOR
  \end{algorithmic}
\end{algorithm}
\setlength{\textfloatsep}{\textfloatsepsave}
%

%TODO Comment on computational complexity of alg.

\parhead{Sparse GP approximation.}
To scale our method to big datasets, we approximate the latent GP $\boldf$ by a {\it sparse Gaussian process} building on {\it inducing points}.
We introduce $M$ inducing points $\bu$ and connect the GP values with the inducing points via the joint prior distribution $p(\boldf, \bu)$ given in \citet{titsias2009variational}.
The introduction of inducing points preserves conditional conjugacy and allows for mini-batch sampling of the data (stochastic variational inference). This scales the algorithm to big datasets and has the computational complexity $\mathcal{O}(M^3)$.
The SVI version of our algorithm only slightly changes the updates that are presented in Algorithm~\ref{alg:VI}. It is deferred to Appendix~\ref{appendix:sparse}.

%\parhead{Convexity of the ELBO and convergence speed.}
%
%When computing the evidence lower bound (ELBO) (see \ref{appendix:ELBO}), the  one can see that this is sum of a convex and a concave function in the variational parameters of $\boldf$.
%We therefore cannot guarantee that our algorithm will converge to a global optima, however the CAVI algorithm ensure that it will reach a local optima.

%TODO Put this back?
%Nonetheless, our algorithm converges with an exponential rate, which we show in the Appendix~\ref{appendix:bound}.

%\paragraph{Connection to a variational local bound (Jaakkola, McKay)\ref{appendix:bound}}
%\com{Need to fill this section ? or put it in related work?}
\vspace{-0.3cm}

\subsubsection{Gibbs sampling}
\label{subsec:Gibbs}

To sample from the exact posterior distribution, a Gibbs sampling scheme alternates between sampling from the complete conditional distributions. In the following we propose a sampling scheme for the distribution family $\pi_\varphi(\omega|c)$ that is automatically constructed given the \gls{PDR} function of the likelihood $\varphi(\cdot)$

The distribution class $\pi_\varphi$ is defined in Eq.~ \ref{eq:invlapomega} and is based on the inverse Laplace transform of $\varphi(\cdot)$.
However there is no general approach to compute the inverse Laplace in closed form \citep{cohen2007numerical}. %TODO: Theo wrote "ill-posed", but this has to be either explained or removed.
%
%We circumvent this issue and propose a sampling approach that only builds on evaluating the (forward) Laplace transform. We apply the method proposed by \citet{ridout2009generating}, %TODO: That's actually not true, so I've deleted it for now.
We circumvent this issue by proposing an algorithm that only evaluates the inverse Laplace transformation point-wise but does not need access to its full analytical form.
We apply the method proposed by \citet{ridout2009generating}, which build on the fact that the \gls{CDF} $F_{\pi_\varphi(\omega|c)}(\cdot)$ can be computed via the inverse Laplace transform of a scaled (forward) Laplace transform,
\begin{align*}
F_{\pi_\varphi(\omega|c)}(x) &= \mathcal{L}^{-1}\left\{\frac{\mathcal{L}\left\{\pi_\varphi(\omega|c)\right\}(s)}{s}\right\}(x)\\
&= \mathcal{L}^{-1}\left\{\frac{\varphi(s+c^2)}{s\varphi(c^2)}\right\}(x).
\end{align*}
To generate samples from $\pi_\varphi(\omega|c)$, we first generate a uniform sample $u\sim\mathcal{U}\left[0,1\right]$ and then push it through the inverse \gls{CDF}, $\omega=F^{-1}_{\pi_\varphi(\omega|c)}(u)$ \citep{devroye1986nonuniform}
%Samples are obtained via the inverse transform sampling : we take a sample $u$ from the uniform distribution $\mathcal{U}\left[0,1\right]$, and we apply the inverse function $F^{-1}_{\pi_\varphi(c)}$.
Finally, to compute the inverse \gls{CDF}, we solve a fixed point problem using the modified Newton-Raphson method described by \citet{ridout2009generating}. We solve the equation $F_{\varphi(c)}(\omega)=u$ by repeatedly setting $\omega \leftarrow \omega - F_{\varphi(c)}(\omega)/\pi_\varphi(\omega|c)$ until reaching convergence. We numerically approximate the (forward) \gls{CDF} $F_{\varphi(c)}(\omega)$ by the cheap trapezoidal method introduced in \citet{abate2000introduction}, which has error guarantees. The cost of this process is negligible against the matrix inversion for sampling $\boldf$.
\label{sec:gibbs}
% by iterating $\omega\leftarrow \omega- F_{\varphi(c)}(\omega)/\pi_\varphi(\omega|c)$ until we reach a convergence criteria $\tau$.
%We evaluate $F_{\varphi(c)}(\omega)$ and $\pi_\varphi(\omega|c)$ by approximating the inverse Laplace transforms integrals with a cheap trapezoidal method introduced in \citet[p268-270]{abate2000introduction} with error guarantees.
All steps are summarized in Algorithm~\ref{alg:Gibbs}.

Note that for some likelihood functions (e.g. the logistic likelihood function), the inverse Laplace transform can be derived analytically and the steps described above can be optimized by using an existing the sampler for the corresponding complete conditional distribution.
%
%\textbf{Analytical Probability Density Function:} For some likelihoods, the inverse Laplace transformation of the \gls{PDR} function $\phi(.)$ can be computed analytically.
%%which inverse Laplace transform can be found analytically via tables or symbolic mathematical software such as Mathematica \citep{Mathematica}.
%For instance, for the Laplace likelihood,  $\varphi(r)=\exp(-\sqrt{r}/\beta)$ the inverse Laplace transformation is analytically given and leads to the distribution $\pi_\varphi(\omega)=\mathcal{IG}(-0.5,1/(4\beta^2))$, where $\mathcal{IG}$ is the inverse-gamma distribution.
%\flotheo{But still, we have no general sampler, right? Maybe we actually can delete this paragraph?}

%\paragraph{Analytical Cumulative Density Function:} Laplace transform can be used to define the cumulative density function $F_\omega$ for a distribution:
%\begin{align*}
%	F_\omega(\omega)=&\mathcal{L}^{-1}\left\{\frac{1}{s}\mathcal{L}\{p(\omega)\}(s)\right\}(\omega) \\
%	=& \mathcal{L}^{-1}\left\{\frac{1}{s}\varphi(s)\right\}(\omega)
%\end{align*}
%When a solution is found, and the inverse function can be found, then a naive sampler can be directly implemented.

\begin{algorithm}
  \caption{Gibbs Sampling}
  \label{alg:Gibbs}
  \begin{algorithmic}
    %	\KwData{$\mathbf{X},\by$}
    	\STATE {\bfseries Input:} Data $(\mathbf{X},\by)$, GP model $p(\by|\boldf)$, kernel $k$ %\boldsymbol{\mu}_0,
    	\STATE {\bfseries Output:} Posterior samples {$\{\boldf^t\} \sim
      p(\boldf\mid\by)$}
  		\FOR{sample index $t=1,2,\dots$,}
  		\STATE \verb|# Sample |$\bomega\sim p(\bomega|\boldf, \by)$:
  		\FOR{$i \in 1:N$}
  		\STATE Compute $c_i=\|h(f_i,y_i)\|_2$
  		\STATE Sample $u_i \sim \mathcal{U}[0,1]$
  		\STATE \verb|# Compute inverse cdf |$\omega_i = F^{-1}_{\pi_\varphi(c_i)}(u_i)$:
      \STATE Initialize $\omega_i > 0$
  		\WHILE{$|\widetilde{F}_{\pi_\varphi(c_i)}(\omega_i) - u_i| > \epsilon$}
  		\STATE  Approximate $\widetilde{F}_{\pi_\varphi}(\omega_i)$, $\widetilde{\pi}_\varphi(\omega_i|c_i)$(see Sec.\ref{sec:gibbs})
  		\STATE $\omega_i \leftarrow \omega_i - \frac{\widetilde{F}_{\pi_\varphi(c_i)}(\omega_i)}{\widetilde{\pi}_\varphi(\omega_i|c_i)}$
  		\ENDWHILE
  		\ENDFOR

        \STATE \verb|# Sample |$\boldf\sim p(\boldf|\bomega, \by)$:
    		\STATE $\boldsymbol{\Sigma} = \left(\diag\left(2\bomega \circ \gamma(\by)\right) + K^{-1} \right)^{-1}$
    		\STATE $\boldsymbol{\mu} = \boldsymbol{\Sigma}\left(K^{-1}\mu_0 + g(\by) + \bomega \circ \beta(\by)\right)$
        \STATE Sample $\boldf^t \sim \mathcal{N}\left(\boldsymbol{\mu},\boldsymbol{\Sigma}\right)$
    \ENDFOR
  \end{algorithmic}
\end{algorithm}
\setlength{\textfloatsep}{\textfloatsepsave}%

 %%%%%%%%%%%%%%%%%%%%%%%%%%%%%%%%%%%
%\section{Additional Stuff}
%\label{sec:theory}
%\glsresetall
%%%%%%%%%%%%%%%%%%%%%%%%%%%%%%%%%%%
%
%\begin{itemize}
%    \item Constructing new super-Gaussian likelihoods
%    \item convergence analysis
%    \item augmentation gap (\flo{we briefly should repeat the derivation of the gap from the x-gpc paper somewhere})
%    \item \flotheo{Connection to Jaakola, and say we are more general.}
%\end{itemize}

 %%%%%%%%%%%%%%%%%%%%%%%%%%%%%%%%%%
\section{Related Work}
\label{sec:related_work}
\glsresetall
%%%%%%%%%%%%%%%%%%%%%%%%%%%%%%%%%%
Inference for non-conjugate likelihoods is not a new topic and there have been many works to deal efficiently with the problem.

\parhead{Scale mixtures of normals.}
The Gaussian scale-mixture formulation is well known in statistics and have been explored more recently by \citet{gneiting1997normal,gneiting1999radial}. \citet{palmer2006variational,articlepalmer2006variational} started to generalize it for a machine learning use but did not explore the probability side of the augmentation.

\parhead{Black-box variational inference.} One of the most popular approach for variational inference in the recent years is to optimize the ELBO for an arbitrary model by computing gradients estimates via sampling or quadrature, e.g. \citet{salimbeni2018natural,mohamed2019monte}.
However these methods do not exploit the structure of the model and can be less efficient.

\parhead{Sampling methods.} Sampling is not a popular method for GP models since $f$ is high-dimensional and the posterior is usually highly correlated \citep{lawrence2009efficient}.
But as for many Bayesian models, Hamiltonian Monte Carlo is a good candidate  \citep{titsias2008markov}.

\parhead{Likelihood approximation.}
\citet{jaakkola2000bayesian} propose a variational approach purely based on optimization, using the partial convexity of the likelihood. Our method recovers their results, but coming from a probabilistic perspective.
We show in Appendix \ref{appendix:bound}, the equivalence with their approach.
\citet{khan2017conjugate} exploit existing partial conjugacy in the model and rely on the assumption that part of the joint posterior can be rewritten as an exponential family.
Their approach is complementary to ours and could be combined for solving more complex models.

\parhead{Use cases of the augmented model.} Different applications of the augmentation technique for specific likelihoods have been explored in multiple papers: \citet{jylanki2011robust} applied the augmentation on the Student-t likelihood with Gaussian Processes. \citet{polson2013bayesian} developed an approach with the logistic likelihood, this work was further expanded by \citet{wenzel2019efficient} to big data.
The augmentation done on the Bayesian Support Vector Machine of \citet{polson2011data} and scaled up by \citet{wenzel2017bayesian}, is similar to our method but is based on a different augmentation approach. 
%It is nonetheless applicable there as well, and present the advantage to not give an improper prior.
Note that our method covers all these cases exactly but do not rely on any manual derivations.
%The scale-mixture development is of course not new and there has been a huge work of Palmer describing likelihoods as "super-gaussian" or "sub-gaussian" and having conditions equivalent to our development.
%However Palmer did not explore the connection with the PDR functions and all the useful properties that they entail.
%There has also been a lot of papers exploring augmentations for specific likelihoods, such as the logistic likelihood \needcite, Bayesian hinge loss \needcite or multi-class likelihoods \needcite

 %%%%%%%%%%%%%%%%%%%%%%%%%%%%%%%%%%
\vspace{-0.5cm}
\section{Experiments}
\vspace{-0.2cm}
\label{sec:experiments}
\setlength{\belowcaptionskip}{-10pt}
\setlength{\abovecaptionskip}{0pt}
\setlength{\floatsep}{0pt}
\setcounter{figure}{2}
\begin{figure*}[h]
	\begin{center}
		\includegraphics[width=0.8\textwidth]{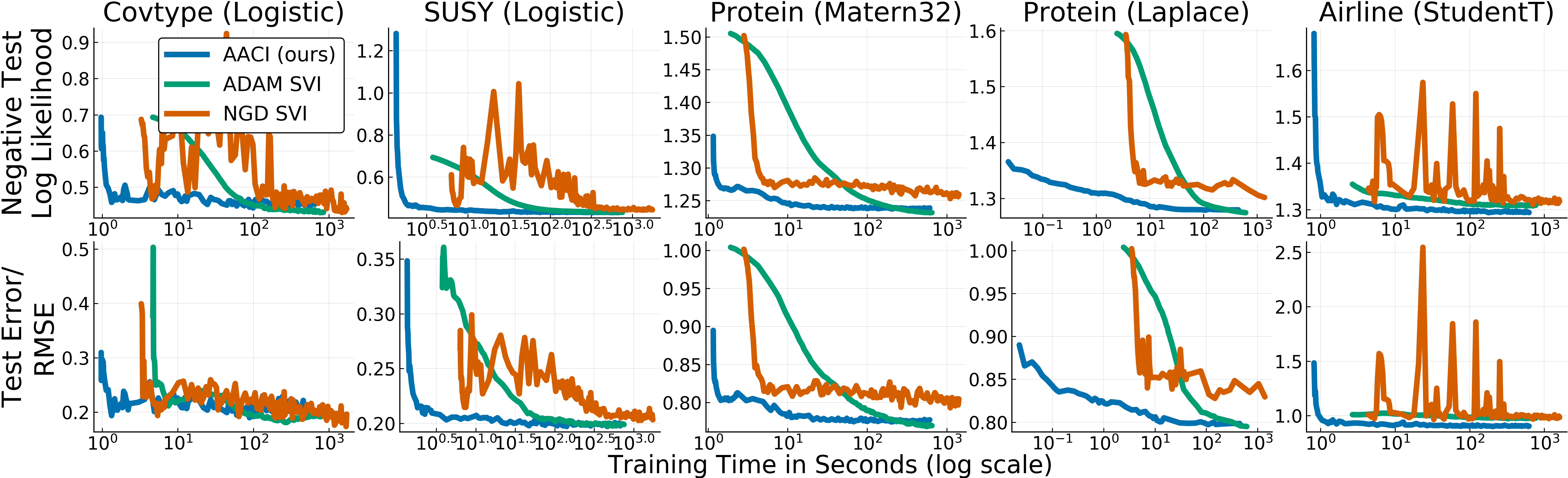}
		\caption{Test negative log-likelihood and test error (classification)/RMSE (regression) as a function of time for different likelihoods.}
		\label{fig:convspeed}
	\end{center}
\end{figure*}
%%%%%%%%%%%%%%%%%%%%%%%%%%%%%%%%%%

In this section we answer the following questions empirically:
\begin{compactitem}
	\item How does the Gibbs sampling scheme compare to other sampling methods?
	\item What is lost in variational inference by approximating an additional variable?
	\item And what is the gain in speed?
\end{compactitem}
We explore four different cases.
We use three regression models with different likelihood functions: a Laplace likelihood, a Student-t likelihood, a new likelihood inspired by the Matern 3/2 kernel \citep{rasmussen2003gaussian} and one classification model with a logistic likelihood.
All the mathematical details of these augmentations are deferred to the Appendix \ref{appendix:explikelihood}. For the two first experiments we use a full GP without inducing points to have a cleaner analysis of the effect of the augmentation.
For all experiments we use a squared exponential kernel with automatic relevance determination: $k(x,x') = \exp(-\sum_{d=1}^D(x_d-x'_d)^2/\theta_d^2)$.
For the two first experiments we use datasets from the UCI repository \citep{Dua:2019} : the Boston housing dataset ($N=506,D=14$) for regression and the Heart dataset ($N=303,D=14$) for classification.
For the last experiment we use the Protein dataset ($N=45730,D=9$) and the Airline dataset ($N=190K,D=7$) for regression and the Covtype dataset ($N=581K,D=54$) and the SUSY dataset ($N=5$M, $D=18$) for classification.
We normalize the input features to mean 0 and variance 1.
\vspace{-0.3cm}
\subsection{Gibbs sampling mixing}
\label{sec:expgibbs}
Our approach leads to a Gibbs sampling algorithm that provides samples from the true posterior of the original model. 
We compare our method (\textit{Gibbs}) with a naive Metropolis-Hasting algorithm (\textit{MH}) and a Hamiltonian Monte Carlo (\textit{HMC}) sampler (where $\epsilon$ and $n_{step}$ are selected via a grid search, see appendix \ref{appendix:fig}) both implemented in Turing.jl \citep{ge2018t}, with a whitening transformation on the kernel matrix for better mixing.
We draw 5 independent chains of 10000 samples for each algorithm.
We compare crucial sampling diagnostics among different models:
%- the mixing time via the Heidelberger test \citep{heidelberger1983simulation}, i.e. the number of samples needed to reach the stationary distribution,\\
we give the autocorrelation between consecutive samples (lag 1) (as well as the autocorrelation plots for all lags in appendix~\ref{appendix:fig}) to estimate the efficient sample size and the chain intercorrelation via the Gelman test (1 is the optimum) \citep{brooks1998general}.
The results are summarized in table \ref{table:sampling}.

\begin{table}
\begin{center}
\begin{tabular}{ccccc}
	\multicolumn{2}{c}{Likelihood/Method} & MH & HMC & \textbf{Gibbs}\\\hline
	\multirow{3}{30pt}{Logistic} & Time/Sample (s) & \textbf{0.001} & 0.041  & 0.01\\
%	& Mix. Time & 1600 & 91 & 32\\
	& Lag 1 & 0.996& 0.53& \textbf{0.11}\\
	& Gelman & 1.38 & \textbf{1.00} & \textbf{1.00}\\\hline
	\multirow{3}{33pt}{\small Student-t}	& Time/Sample (s) & \textbf{0.003} & 0.573 & 0.028 \\
	& Lag 1 & 1.0 & 0.857& \textbf{0.04}\\
%	& Mix. Time & 1276 & 594 & 96\\
	& Gelman & 1.51 & 1.00 & \textbf{1.00}\\\hline
	\multirow{3}{30pt}{Laplace}	& Time/Sample (s) & \textbf{0.002} & 0.082 & 0.028\\
	& Lag 1 & 0.995& 0.931& \textbf{0.26} \\
%	& Mix. Time & 1752 & 1873 & 132\\
	& Gelman & 1.44 & 1.01 & \textbf{1.00}\\\hline
	\multirow{3}{30pt}{Matern 3/2}	& Time/Sample (s) & \textbf{0.005} & 0.15 & 0.029\\
	& Lag 1 & 0.997& 0.995& \textbf{0.05}\\
%	& Mix. Time & 562 & 1061 & 125\\
	& Gelman & 1.59 & 1.10 & \textbf{1.00}\\\hline
\end{tabular}
\caption{Sampling time and diagnostics of Gibbs Sampling, naive Metropolis-Hastings and Hamiltonian Monte-Carlo. %	The mixing time corresponds to the number of samples needed to reach the stationary distribution.
The Gelman test indicates the inter-chain correlation and should be close to 1.}
\label{table:sampling}
\end{center}
\end{table}
\setcounter{figure}{1}
\begin{figure}[h]
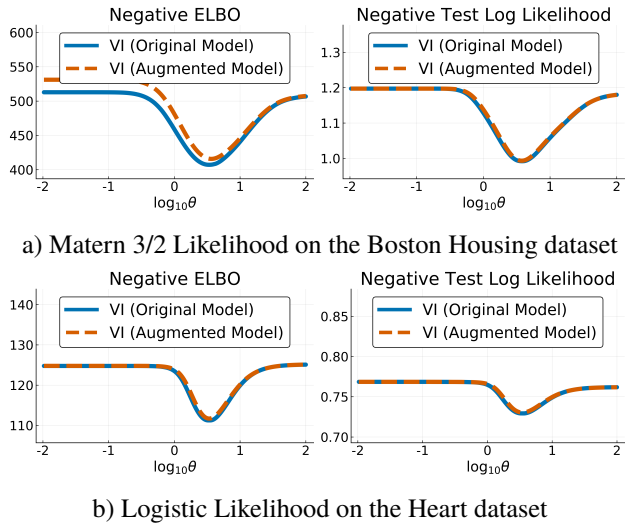

	\includegraphics[width=0.49\columnwidth]{\plotdir{{part_2/ELBO_likelihood=Laplace_variance=0.1}.png}} \includegraphics[width=0.49\columnwidth]{\plotdir{{part_2/NLL_likelihood=Matern32_variance=0.1}.png}}
	\begin{center}
		a)	Matern 3/2 Likelihood on the Boston Housing dataset
	\end{center}
	\includegraphics[width=0.49\columnwidth]{\plotdir{{part_2/ELBO_likelihood=Logistic_variance=0.1}.png}} \includegraphics[width=0.49\columnwidth]{\plotdir{{part_2/NLL_likelihood=Logistic_variance=0.1}.png}}
	\begin{center}
		b)	Logistic Likelihood on the Heart dataset
	\end{center}
	\caption{Converged negative ELBO and averaged negative log-likelihood on a held-out dataset in function of the kernel lengthscale, training VI with and without augmentation.}
	\label{fig:elbovi}
\end{figure}
%\begin{figure}
%	\includegraphics[width=0.49\columnwidth]{plots/part_1/Logistic/lag_plot.png}
%	\includegraphics[width=0.49\columnwidth]{plots/part_1/StudentT/lag_plot.png}
%	\includegraphics[width=0.49\columnwidth]{plots/part_1/Laplace/lag_plot.png}
%	\includegraphics[width=0.49\columnwidth]{plots/part_1/Matern32/lag_plot.png}
%	\caption{Auto-correlation for different lag (iterations between samples).}
%\end{figure}
We find that our method has a very low intrachain correlation leading to a high sample efficiency, as well as a low interchain correlation while still being faster than the HMC algorithm.
It is even more evident for heavy-tailed likelihood like Student-T or Laplace where HMC can be of more trouble \citep{betancourt2017conceptual}.
Our approach is limited by the $\mathcal{O}(N^3)$ complexity for each sample.

\subsection{Augmentation gap}
\label{sec:auggap}
To investigate the effect of augmenting the model when using variational inference, we train the original model using gradient descent and the augmented model until convergence.
While we fix the kernel variance at 0.1, we vary the lengthscale $\theta$ from $10^{-2}$ to $10^{2}$.
We compare the converged ELBOs as well as the predictive performance on held-out test set.
The results for the matern 3/2 and logistic are shown on figure~\ref{fig:elbovi}, the other likelihoods are show in the appendix~\ref{appendix:fig}.
For both shown likelihoods, there is a visible ELBO gap between the augmented model and the original model.
However the predictive performance is marginally the same for both models.% although the augmented model is marginally worse.
We can conclude that a potential difference in ELBO values does not affect the prediction performance. %\cite{jaakkola2000bayesian}.

\vspace{-0.3cm}

\subsection{Convergence speed}
\label{sec:convspeed}
%One of the main objectives is to have a VI both faster and more stable compared to the original model.
To scale our model to large datasets, we use the inducing points technique of \citet{titsias2009variational} and we use the stochastic gradient descent approach of \citet{hoffman2013stochastic}.
We compare our variational approach (Algorithm \ref{alg:VI}) to using natural gradient descent, \citep{salimbeni2018natural} and ADAM \citep{hensman2015scalable} both implemented in GPFlow \citep{matthews2017gpflow}.
For all methods we use 200 inducing points determined by $k$-means++ \citep{arthur2007k}, minibatches of size 100 and we train the kernel hyperparameters using ADAM \citep{kingma2014adam}, (the inducing points locations are fixed).
%We do not consider the method of \citet{khan2017conjugate} since it is not a fully automated method.\com{to improve}
We show the predictive performance in function of the training time for multiple likelihoods on figure \ref{fig:convspeed}.

Our method is up to two orders of magnitude faster than the state of the art.
Moreover, we find that the optimization in our method is more stable (smooth decrease of the loss.
\vspace{-0.3cm}
 %%%%%%%%%%%%%%%%%%%%%%%%%%%%%%%%%%
\section{Conclusion}
\label{sec:conclusion}
%%%%%%%%%%%%%%%%%%%%%%%%%%%%%%%%%%
\vspace{-0.4cm}
We proposed a new efficient inference method for GP models that have a super-Gaussian likelihood. 
Our method builds on an auxiliary variable augmentation that renders the model conditionally conjugate. 
We showed that in the augmented model, variational inference is up to two orders of magnitude faster and more stable than the state of the art. 
For small dataset, we proposed a Gibbs sampler that outperforms Hamiltonian Monte Carlo sampling.
Previous methods that build on auxiliary variable augmentations \citep[e.g.][]{wenzel2019efficient} manually derived the augmentation and inference methods, whereas in our approach the whole procedure is fully automated and works for much more general class of models.
Future work may aim on extending our approach to more general models by automatically constructing \emph{hierarchical augmentations} inspired by \citet{galy2019multi} or \citet{donner2018efficient}.
%which could by inspired by the (manually) constructed hierarchical augmentations proposed by \citet{galy2019multi} and \citet{donner2018efficient}.

%\subsubsection*{Acknowledgements}
%Thank you everyone, you are all beautiful people.
\bibliographystyle{apalike}
\bibliography{autoconjugate}

\onecolumn
\appendix
\newpage
\section{Appendix}

\subsection{Proof of theorem~\ref{thm:fullcond}}
\label{appendix:fullcond}
Theorem~\ref{thm:fullcond} states:
\begin{theorem*}
	The \textbf{complete conditional distributions} of the augmented model presented in Section~\ref{subsec:augmentation} are given by
	\begin{align*}
	\begin{aligned}
	p(\omega_i|f_i,y_i) =& \pi_{\varphi}\left(\omega_i| \|h(f_i,y_i)\|_2\right),\\
	p(\boldf|\by,\bomega) =& \mathcal{N}\left(\boldf| \boldsymbol{\mu} ,\boldsymbol{\Sigma}\right),
	\end{aligned}
	\end{align*}
	where $\boldsymbol{\Sigma} = \left(\diag\left(2\bomega\circ \gamma(\by)\right) + K^{-1}\right)^{-1}$ and $\boldsymbol{\mu} = \boldsymbol{\Sigma}\left(g(\by)+\bomega\circ \beta(\by) + K^{-1}\boldsymbol{\mu}_0\right)$, $\circ$ denotes the Hadamard product and the function $h(\cdot)$ is given by the form of likelihood (see  Eq.\ref{eq:h}).
\end{theorem*}
\parhead{Proof:}
For the full conditional on $\boldf$:
\begin{align*}
p(\boldf|\by,\bomega) \propto& p(\by|\boldf,\bomega)p(\boldf)\\
\propto& \exp\left[g(\by)^\top \boldf + (\beta(\by)\circ \bomega)^\top\boldf -\boldf^\top\diag(\gamma(\by)\circ \bomega)\boldf - \half \boldf^\top K^{-1}\boldf   \right]\\
\propto& \exp\left[\left(g(\by)+\beta(\by)\circ \bomega\right)^\top \boldf  -\boldf^\top\left[\diag(\gamma(\by)\circ \bomega)+\half K^{-1}\right]\boldf  \right].
\end{align*}
We get immediately a multivariate normal distribution with $-\half \Sigma^{-1} = -\diag(\gamma(\by)\circ \bomega)+\half K^{-1}$ and $\Sigma^{-1}\mu = g(\by) + (\beta(\by)\circ \bomega)$. Which corresponds to the result shown in equation (\ref{eq:fullconditionals}).

For the augmented variable $\omega_i$:
\begin{align*}
p(\omega_i|y_i,f_i) \propto& p(y_i|f_i,\omega_i)p(\omega_i)\\
\propto & \exp\left( -\|h(y_i,f_i)\|_2^2\omega_i\right)\pi_\varphi(\omega_i|0)\\
=&\pi_\varphi(\omega_i|\|h(y_i,f_i\|_2).
\end{align*}
Note that the equation~\ref{eq:scale-mixture} gives the normalization constant directly $\varphi(\|h(y_i,f_i)\|_2^2)$ directly.
QED.

\subsection{Computation of the moments and cumulants for the augmentation variable}
\label{appendix:moment}
Given the general class of distribution $\pi_\varphi(\omega|c)$ described in Section~\ref{subsec:augmentation}, moments and cumulants can be easily computed:
The $k$-th moment of a distribution can be computed by taking the $k$-th derivative of the moment generating function (equivalent to  a negative Laplace transform) at $t=0$. For example for the first moment:
\begin{align*}
	\expec{\pi_\varphi(\omega|c)}{\omega} =& \left.\frac{d\laplace{\pi_\varphi(\omega|c)}{-t}}{dt}\right|_{t=0}\\ =&\left.\frac{d}{dt}\left[\mathcal{L}\left[\frac{e^{-c^2\omega}\pi_\varphi(\omega|0)}{\varphi(c^2)}\right](-t)\right]\right|_{t=0}\\
	=& -\frac{1}{\varphi(c^2)}\left.\frac{d}{dt}\left[\mathcal{L}\left[\pi_\varphi(\omega|b,0)\right](t+c^2)\right]\right|_{t=0}\\
	=& \left.-\frac{1}{\varphi(c^2)}\frac{d\varphi\left(t+c^2\right)}{dt}\right|_{t=0}\\
	=& - \left.\frac{d\log\varphi(t)}{dt}\right|_{t=c^2}\\
	=& - \frac{\varphi'(c^2)}{\varphi(c^2)} = \overline{\omega}
\end{align*}

More generally the $k$-th moment $m_k$ is defined as :
\begin{align*}
	m_k =& (-1)^k\frac{1}{\varphi(c^2)}\left.\frac{d^k\varphi(t)}{dt^k}\right|_{c^2}
\end{align*}
And the cumulants $\kappa_k$ are computed using the cumulant generating function (log of the moment generating function)
\begin{align*}
	\kappa_k =& (-1)^k\left.\frac{d^k\log \varphi(t)}{dt^k}\right|_{t=c^2}\\
\end{align*}

\subsection{Algorithm for the sparse case}
\label{appendix:sparse}
\begin{algorithm}
	\caption{Augmented Stochastic Variational Inference}
	\label{alg:sparseVI}
	\begin{algorithmic}
		%	\KwData{$\mathbf{X},\by$}
		\STATE {\bfseries Input:} Data $(\mathbf{X},\by)$, GP model $p(\by|\boldf,\boldsymbol{u})$, kernel $k$ %\boldsymbol{\mu}_0,
		\STATE {\bfseries Output:} Approximate posterior {$q(\boldsymbol{u}) =
			\mathcal{N}(\boldsymbol{u}|\boldsymbol{m},\boldsymbol{S})$}
		\STATE Find inducing points inputs $Z$ via $k$-means
		\STATE Compute kernel matrices : $K_{Z}$, $\kappa = K_{XZ}K_Z^{-1}$
		\FOR{iteration $t=1,2,\dots$,}
		\STATE \verb|# Local updates:|
		\STATE Sample minibatch $\mathcal{B} \subseteq \{1,\dots,n\}$
		\FOR{$i \in \mathcal{B}$}
		\STATE $c_i=\sqrt{\expec{q(f)}{h(f_i,y_i)^2}}$
		\STATE $\overline{\omega}_i = \expec{q(\omega_i)}{\omega_i} =  -\varphi'(c_i^2)/\varphi(c_i^2)$
		\ENDFOR
		\STATE \verb|# Natural gradient updates (CAVI):|
		\STATE $\widetilde{\boldsymbol{S}} = \left(\kappa^\top\diag\left(2\overline{\bomega} \circ \gamma(\by)\right)\kappa + K_Z^{-1} \right)^{-1}$
		\STATE $\widetilde{\boldsymbol{m}} = \widetilde{\boldsymbol{S}}\left(K_Z^{-1}\mu_0 + \kappa^\top \left(g(\by) + \overline{\bomega} \circ \beta(\by)\right)\right)$
		\STATE $\{\boldsymbol{m}, \boldsymbol{S}\} \leftarrow  (1-\rho^{(t)})\{\boldsymbol{m}, \boldsymbol{S}\} +  \rho^{(t)}\{\widetilde{ \boldsymbol{m}}, \widetilde{\boldsymbol{S}}\}$
		\ENDFOR
	\end{algorithmic}
\end{algorithm}

$\rho^{(t)}$ is an arbitrary learning rate respecting the Robbins-Monroe condition.

\subsection{ELBO Analysis}
\label{appendix:ELBO}
\subsubsection{Full ELBO}
\begin{align*}
\begin{split}\ELBO =& \sum_{i=1}^N\expec{q(f_i,\omega_i)}{\log p(y_i|f_i,\omega_i)}\\
&- \KL[q(f)||p(f)] - \sum_{i=1}^N\KL[q(\omega_i)||p(\omega_i)]
\end{split}
\end{align*}
%%% Expec log likelihood
\begin{align*}
\expec{q}{\log p(y_i|f_i,\omega_i,\theta)} =&\log C(\theta) + g(y_i,\theta)\expec{q(f)}{f} - \expec{q(f)}{h(f_i,y_i)^2}\expec{q(\omega_i)}{\omega_i}\\
=& \log C(\theta) + g(y_i,\theta)m_i-\left(\alpha(y_i)-\beta(y_i)m_i+\gamma(y_i)\left(m_i^2 + S_{ii}\right)\right)\overline{\omega}_i\nonumber\\
\KL[q(f)||p(f)]=& \frac{1}{2}\left[\log \frac{|K|}{|\boldsymbol{S}|}-N+\tr(K^{-1}\boldsymbol{S})+(\boldsymbol{\mu}_0-\boldsymbol{m})^\top K^{-1}(\boldsymbol{\mu}_0-\boldsymbol{m})\right]\nonumber\\
\KL[q(\omega_i)||p(\omega_i)] =& -\expec{q(\omega_i)}{c_i^2\omega_i} - \log \varphi(c_i^2) = -c_i^2\overline{\omega}_i-\log \varphi(c_i^2)\nonumber
\end{align*}
Note that we can take the derivatives of the ELBO and set them to 0 to recover exactly the updates in algorithm \ref{alg:VI}.

%\subsection{Proof of convexity}
%\label{appendix:convexity}
%Taking the second derivative of (\ref{eq:ELBO}) given $\mu$ and $\Sigma$ and by fixing $c_i^2$ one gets:
%\begin{align*}
%\frac{d^2\ELBO}{d\mu d\mu^\top} =& -2\diag(\gamma\circ m) - K^{-1} < 0\\
%\frac{d^2ELBO}{d\Sigma^2}=& -K^{-1}\otimes K^{-1} <0
%\end{align*}
%as $K^{-1}$ is by definition positive definite (and the Kroenecker product $\otimes$ conserves it), and $m_i>0\;~\forall i$.
%Therefore the (negative) ELBO is (convex) concave in $\mu$ and $\Sigma$ indicating that there exists a unique global optimum for the global parameters.

\subsubsection{Analysis of the optima}

By setting $c_i^2$ as a function of $\boldsymbol{m}$ and $\boldsymbol{S}$ (and setting $\boldsymbol{\mu}_0$ to 0 for simplicity) we can get an ELBO only depending of the variational parameters of $\boldf$.
\begin{align*}
\ELBO(\boldsymbol{m},\boldsymbol{S}) = C + g^\top\boldsymbol{m} + \frac{1}{2}\left(\underbrace{\log |\boldsymbol{S}| - \tr(K^{-1}\boldsymbol{S})- \boldsymbol{m}^\top K^{-1}\boldsymbol{m}}_{\ELBO_1}\right) + \sum_i\underbrace{\log \varphi (m_i^2+S_{ii})}_{\ELBO_2}\\
%=C + g^\top\mu - \frac{1}{2}\left(-N\log2 + \log |\eta_2| - \frac{1}{2}\tr(K^{-1}\eta_2^{-1})-2\mu_0K^{-1}\mu + \mu^\top K^{-1}\mu\right) + \sum_i\log \varphi (\mu_i^2+\Sigma_{ii})
\end{align*}

It is easy to show that $\ELBO_1$ is jointly concave in  $\boldsymbol{m}$ and $\boldsymbol{S}$ with a short matrix analysis. However $\ELBO_2$ is more complex : $m_i^2+S_{ii}$ is jointly convex in $\boldsymbol{m}$ and $\boldsymbol{S}$, $\phi(r)$ is by definition convex as well, however $\phi(m_i^2+S_{ii})$ is neither jointly convex or concave in $\boldsymbol{m}$ and $\boldsymbol{S}$.
It is therefore impossible to guarantee that there is a global optima, however the CAVI updates guarantee us a local optima.

\subsubsection{ELBO Gap}

For a fixed $q(f)$ we can compare the ELBO of the original model $\mathcal{L}_{std}(q(f))$ and the augmented model $\mathcal{L}_{aug}(q(f)q(\omega))$. It is then straightforward to compute the difference between the two :

\begin{align*}
\Delta \mathcal{L}=&\mathcal{L}_{std}(q(f)) - \mathcal{L}_{aug}(q(f)q(\omega))\\
=& \expec{q(f)}{\log p(y,f)-\log q(f)-\expec{q(\omega)}{p(y,f,\omega)-\log q(f)q(\omega)}}\\
=& \expec{q(f)q(\omega)}{-\log \frac{p(y,f,\omega)}{p(y,f)} + \log q(\omega)}\\
=& \expec{q(f)q(\omega)}{-\log p(\omega|y,f)+\log q(\omega)}\\
=& \expec{q(\omega)}{\log q(\omega) - \expec{q(f)}{\log p(\omega|y,f)}}\\
%&= \KL\left(q(\omega)||\frac{1}{Z}\exp\left(\expec{q(f)}{\log p(\omega|y,f)}\right)\right)\\
=& -c^2\expec{q(\omega)}{\omega}+\expec{q(\omega)}{\log \PG(\omega|1,0)}-\log\varphi(c^2)\\
&+\expec{q(f)}{f^2}\expec{q(\omega)}{\omega}-\expec{q(\omega)}{\log \PG(\omega|1,0)}+\expec{q(f)}{\log\varphi(f^2)}\\
=& -c^2 m - \log \varphi(c^2) + \expec{q(f)}{f^2}m + \expec{q(f)}{\log \varphi(f^2)}\\
\end{align*}
Replacing with the optimal $q^*(\omega)=\frac{e^{-c^2\omega}p(\omega)}{\varphi(c^2)}$ with $c^2=\expec{q(f)}{f^2}$
\begin{align*}
\Delta\mathcal{L}^* =& - \log \varphi(c^2) + \expec{q(f)}{\log \varphi(f^2)}
\end{align*}

\subsubsection{Sparse ELBO}
When using the inducing points approach the ELBO becomes:
\begin{align*}
\begin{split}\ELBO =& \sum_{i=1}^N\expec{q(f_i,u_i,\omega_i)}{\log p(y_i|f_i,u_i,\omega_i)} \\
&- \KL[q(u)||p(u)] - \sum_{i=1}^N\KL[q(\omega_i)||p(\omega_i)]
\end{split}
\end{align*}
%%% Expec log likelihood
\begin{align*}
\expec{q}{\log p(y_i|f_i,\omega_i,\theta)} =&\log C(\theta) + g(y_i,\theta)\expec{q(f,u)}{f} - \expec{q(f,u)}{h(f_i,y_i)^2}\expec{q(\omega_i)}{\omega_i}\\
=& \log C(\theta) + g(y_i,\theta)(\kappa^\top \boldsymbol{m})_i-\left(\alpha(y_i)-\beta(y_i)(\kappa^\top \boldsymbol{m})_i+\gamma(y_i)\left((\kappa^\top\boldsymbol{m})_i^2 + (\kappa^\top \boldsymbol{S}\kappa)_{ii}\right)\right)\overline{\omega}_i\nonumber\\
\KL[q(f)||p(f)]=& \frac{1}{2}\left[\log \frac{|K|}{|\boldsymbol{S}|}-N+\tr(K^{-1}\boldsymbol{S})+(\boldsymbol{\mu}_0-\boldsymbol{m})^\top K^{-1}(\boldsymbol{\mu}_0-\boldsymbol{m})\right]\nonumber\\
\KL[q(\omega_i)||p(\omega_i)] =& -\expec{q(\omega_i)}{c_i^2\omega_i} - \log \varphi(c_i^2) = -c_i^2\overline{\omega}_i-\log \varphi(c_i^2)\nonumber
\end{align*}

\subsection{Proof of equivalence between Jaakkola bound and data augmentation}
\label{appendix:bound}
\citet{jaakkola2000bayesian} proposed an approach purely based on optimization. They are assuming $\log p(y|f)$ contains a part convex in $f^2$: ${\log p(y|f)=\log p_{convex}(f) + \log p_{non-convex}(f,y)}$. Using convexity properties they are creating a bound with a Taylor expansion to the first order around an additional variable $c^2$:
\begin{align*}
\log p_c(f) \geq \log p_c(c) + \frac{d\log p_c(c)}{dc^2}(f^2-c^2)
\end{align*}
Putting it back in the full ELBO, they are now getting a quadratic part in $f$, analytically differentiable, and they just need to optimize the additional variables $\{c_i\}$.
\citet{merkle2014completely} shows that any completely monotone function is log-convex, i.e. $\log \varphi(r)$ is convex.
Therefore we can replace $\log p_c(c)$ by $\log \varphi(r)$ to recover our model in the context of variational inference.
Note that the converse does not hold, therefore the complete monotonicity is a stronger assumption.

\subsection{Likelihoods used for the experiments}
\label{appendix:explikelihood}
We detail all likelihoods used for the experiments and their formulation as in equation (\ref{eq:likelihood}).

\textbf{Laplace Likelihood :}
${\text{Laplace}(y|f,\beta) = \frac{1}{2\beta}\exp\left(-\frac{|f-y|}{\beta}\right)}$\\
\textbf{Logistic Likelihood :} $p(y|f) = \sigma(yf) = \frac{e^{yf/2}}{2\cosh(|f|/2)}$\\
\textbf{Student-T Likelihood :} $p(y|f) = \frac{\Gamma((\nu+1)/2)}{\Gamma(\nu/2)\sqrt{\pi\nu}}\left(1+\frac{(y-f)^2}{\nu}\right)^{-(\nu+1)/2}$\\
\textbf{Matern 3/2 Likelihood :} $p(y|f) = \frac{4\rho}{\sqrt{3}}\left(1+\frac{\sqrt{3(y-f)^2}}{\rho}\right)\exp\left(-\frac{\sqrt{3(y-f)^2}}{\rho}\right)$\\
%\textbf{Student-T Prior for GP :} $p(f) = \frac{\Gamma\left((\nu+N)/2\right)}{\Gamma(\nu/2)\nu^{N/2}\pi^{N/2}|K|^{1/2}}\left[1+\frac{1}{\nu}(f-\mu_0)^\top K^{-1} (f-\mu_0)\right]^{-(\nu+N)/2}$
\begin{center}
\begin{tabular}{l|c|c|c|c|c|c|c}
Likelihood & $C(\theta)$ & $g(y,\theta)$ & $||h(y,f,\theta)^2||_2^2$ & $\alpha(y)$ & $\beta(y)$ & $\gamma(y)$ &$\varphi(r)$ \\\hline
Laplace & $(2\beta)^{-1}$ & $0$ & $(y-f)^2$ & $y^2$ & $2y$ &$1$ & $e^{-\sqrt{r}/\beta}$\\
Logistic & $2^{-1}$ & $y/2$ & $f^2$ & $0$ & $0$ & $1$ & $\cosh^{-1}(\sqrt{r}/2)$\\
Student-T & $\Gamma((\nu+1)/2)/(\Gamma(\nu)\sqrt{\pi\nu})$ & $0$ & $(y-f)^2$ & $y^2$ & $2y$ &$1$ & $(1+\frac{r}{\nu})^{-(\nu+1)/2}$ \\
Matern 3/2 & $4\rho/\sqrt{3}$ & $0$ & $(y-f)^2$ & $y^2$ & $2y$ & $1$ & $(1+\frac{\sqrt{3r}}{\rho})e^{-\sqrt{3r}/\rho}$\\
%Multivariate-T & $C(\nu,N,K)$ & $0$ & $||L^{-1}(f-\mu_0)||_2^2$ & $(1+r/\nu)^{-(\nu+N)/2}$
\end{tabular}
\end{center}
\newpage

\subsection{Extra figures}
\label{appendix:fig}
\subsubsection{Autocorrelation plots}
\setcounter{figure}{3}
\begin{figure}[h]
	\begin{center}	
	\includegraphics[width=0.32\textwidth]{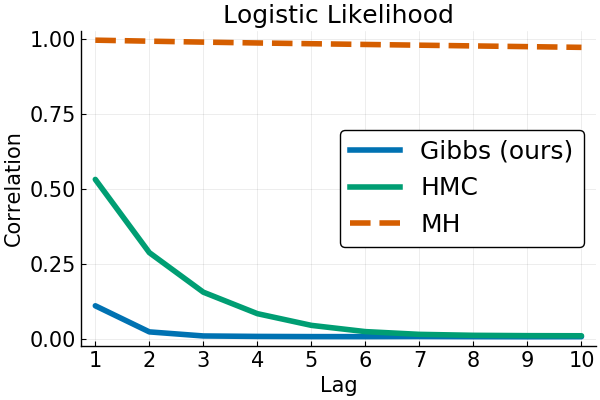}
	\includegraphics[width=0.32\textwidth]{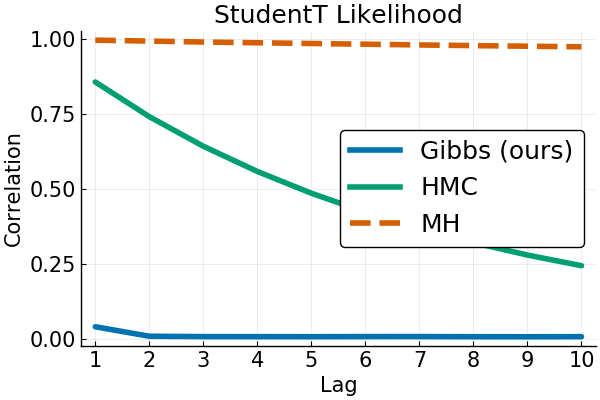}
	
	\includegraphics[width=0.32\textwidth]{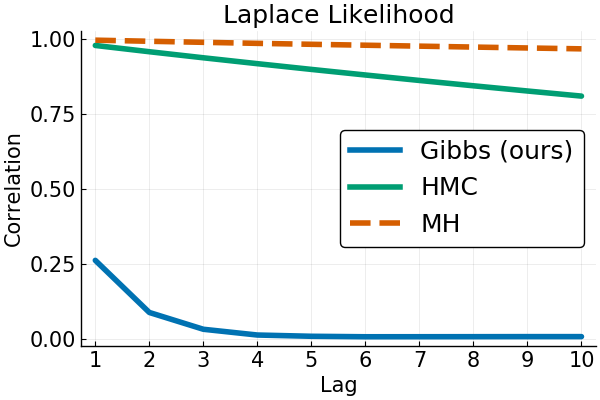}
	\includegraphics[width=0.32\textwidth]{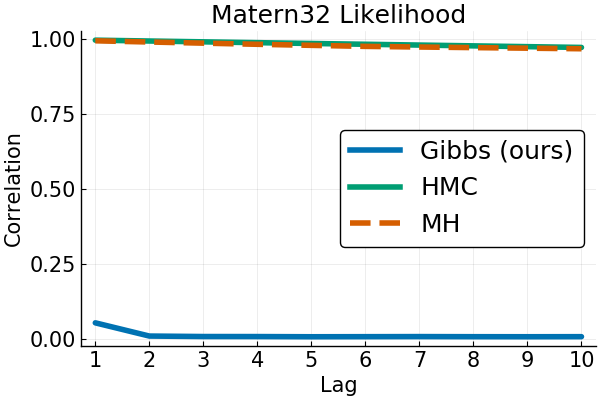}
	\end{center}
	\caption{Auto-correlation plots for differents with lags from 1 to 10}
\end{figure}
\newpage
\subsubsection{HMC Results}

\begin{table}[h]
	\centering
	\begin{tabular}{|c|ccccc|}
		\hline 
		$\epsilon$/$n_{step}$ & & 1 & 2 & 5 & 10 \\ 
		\hline 
		\multirow{3}{30pt}{0.01} & Time/Sample (s)  & 0.037 & 0.045 & 0.077 & 0.133  \\ 
		 & Lag 1 &  0.999 & 0.993 & 0.978 & 0.963\\
		 & Gelman & 3.14 & 1.02 & 1.00 & 2.05\\
		\hline 
		\multirow{3}{30pt}{0.05} & Time/Sample (s)  & 0.036 & 0.046 & 0.080 & 0.12 \\ 
& Lag 1 & 0.999 & 0.998& \textbf{0.931} & 0.948 \\
& Gelman & 1.72 & 1.18& 1.01 & 3.25 \\
\hline 
		\multirow{3}{30pt}{0.1} & Time/Sample (s)  & 0.033 & 0.042 & 0.073 & 0.13 \\ 
& Lag 1 & 0.997 & 0.996& 0.998 & 0.994\\
& Gelman & 1.11 & 1.04& 1.27& 2.71\\
\hline
	\end{tabular} 
	\caption{HMC results for the Laplace likelihood}
\end{table}
\begin{table}[h]
	\centering
	\begin{tabular}{|c|ccccc|}
		\hline 
		$\epsilon$/$n_{step}$ & & 1 & 2 & 5 & 10 \\ 
		\hline 
		\multirow{3}{30pt}{0.01} & Time/Sample (s)  & 0.675 & 0.110 & 0.177 & 0.251  \\ 
		& Lag 1 &  0.999 & 0.999 & 0.997 & 0.993\\
		& Gelman & 3.14 & 1.74 & 1.11 & 1.02\\
		\hline 
		\multirow{3}{30pt}{0.05} & Time/Sample (s)  & 0.148 & 0.192 & 0.336 & 0.573 \\ 
		& Lag 1 & 0.997 & 0.993 & 0.962 & \textbf{0.857} \\
		& Gelman & 1.10 & 1.02& 1.00 & 1.00 \\
		\hline 
		\multirow{3}{30pt}{0.1} & Time/Sample (s)  & 0.142 & 0.193 & 0.337 & NA \\ 
		& Lag 1 & 0.993 & 0.976& 0.864 & NA\\
		& Gelman & 1.03 & 1.01& 1.00 & NA\\
		\hline
	\end{tabular} 
	\caption{HMC results for the Student-T likelihood}
\end{table}
\begin{table}[h]
	\centering
	\begin{tabular}{|c|ccccc|}
		\hline 
		$\epsilon$/$n_{step}$ & & 1 & 2 & 5 & 10 \\ 
		\hline 
		\multirow{3}{30pt}{0.01} & Time/Sample (s)  & 0.009 & 0.013 & 0.021 & 0.041  \\ 
		& Lag 1 &  0.999 & 0.999 & 0.998 & 0.994\\
		& Gelman & 3.19 & 1.68 & 1.12 & 1.02\\
		\hline 
		\multirow{3}{30pt}{0.05} & Time/Sample (s)  & 0.011 & 0.014 & 0.025 & 0.41 \\ 
		& Lag 1 & 0.998 & 0.994& 0.968 & 0.871 \\
		& Gelman & 1.11 & 1.03& 1.00 & 1.00 \\
		\hline 
		\multirow{3}{30pt}{0.1} & Time/Sample (s)  & 0.011 & 0.014 & 0.024 & 0.048 \\ 
		& Lag 1 & 0.994 & 0.979& 0.875 & \textbf{0.532}\\
		& Gelman & 1.02 & 1.01& 1.00& 1.00\\
		\hline
	\end{tabular} 
	\caption{HMC Results for the Logistic likelihood}
\end{table}
\newpage
\subsubsection{ELBO difference}

\begin{figure}[h]
	\begin{center}
	\includegraphics[width=0.32\columnwidth]{\plotdir{{part_2/ELBO_likelihood=StudentT_variance=0.1}.png}} \includegraphics[width=0.32\columnwidth]{\plotdir{{part_2/NLL_likelihood=StudentT_variance=0.1}.png}}\\
		a)	Student-T likelihood on the Boston Housing dataset\\
	\includegraphics[width=0.32\columnwidth]{\plotdir{{part_2/ELBO_likelihood=Laplace_variance=0.1}.png}} \includegraphics[width=0.32\columnwidth]{\plotdir{{part_2/NLL_likelihood=Laplace_variance=0.1}.png}}\\
		b)	Laplace likelihood on the Boston Housing dataset
\end{center}
	\caption{Converged negative ELBO and averaged negative log-likelihood on a held-out dataset in function of the RBF kernel lengthscale, training VI with and without augmentation.}
\end{figure}
\newpage
\subsubsection{Convergence speed}

\begin{figure}[h]
	\begin{center}
	\includegraphics[width=0.4\textwidth]{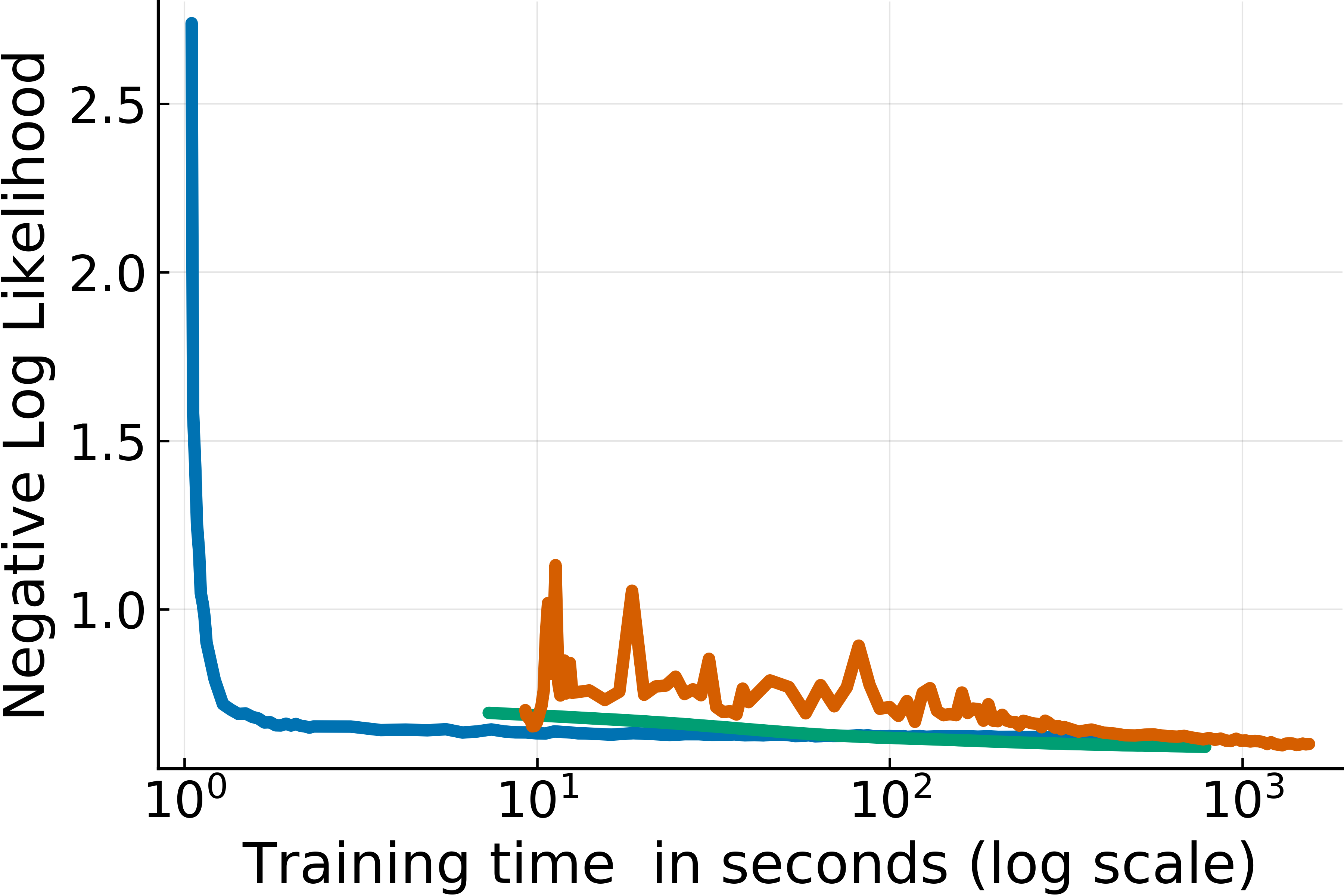}
	\includegraphics[width=0.4\textwidth]{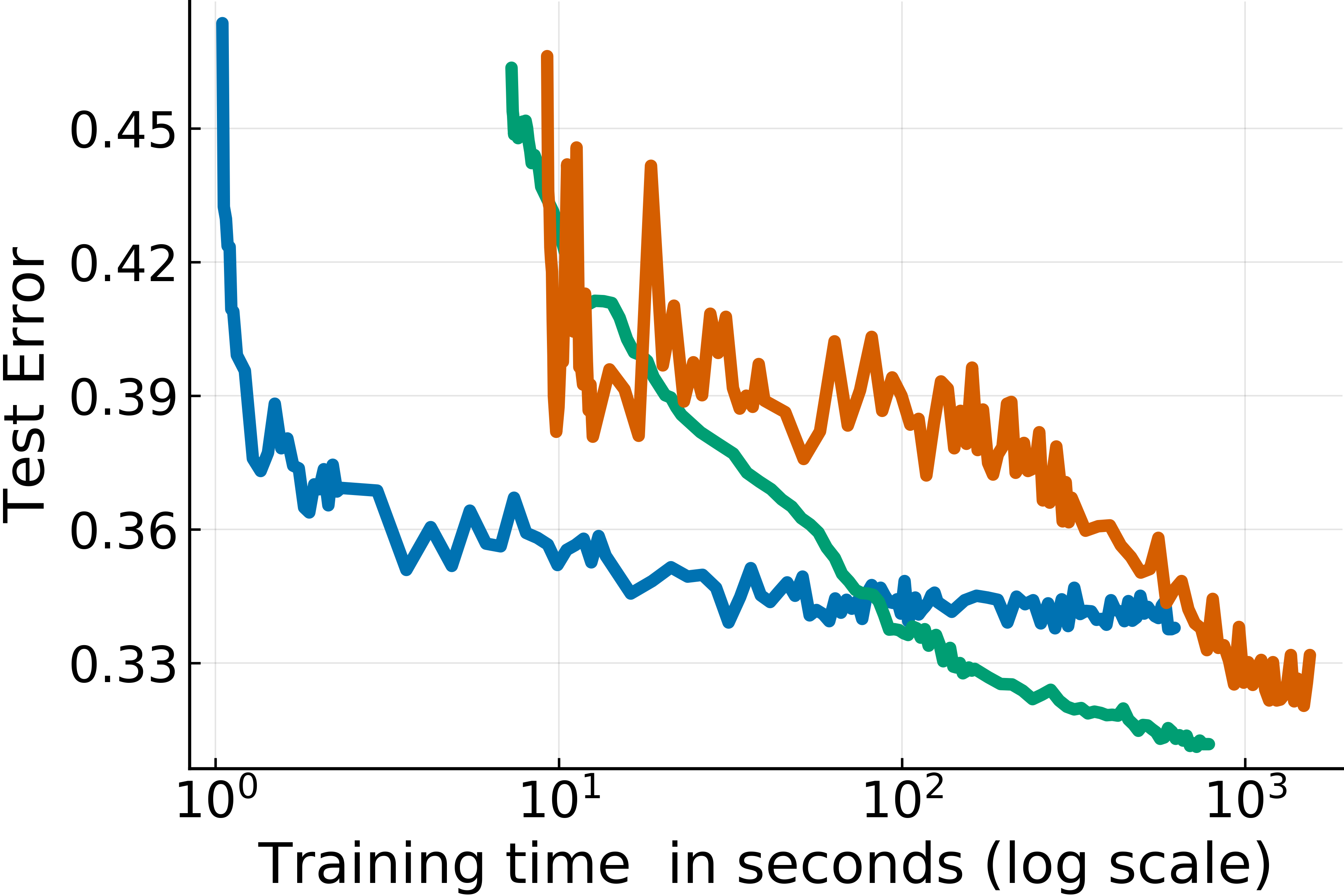}\\
	a) Logistic likelihood on the HIGGS dataset\\
	\includegraphics[width=0.4\textwidth]{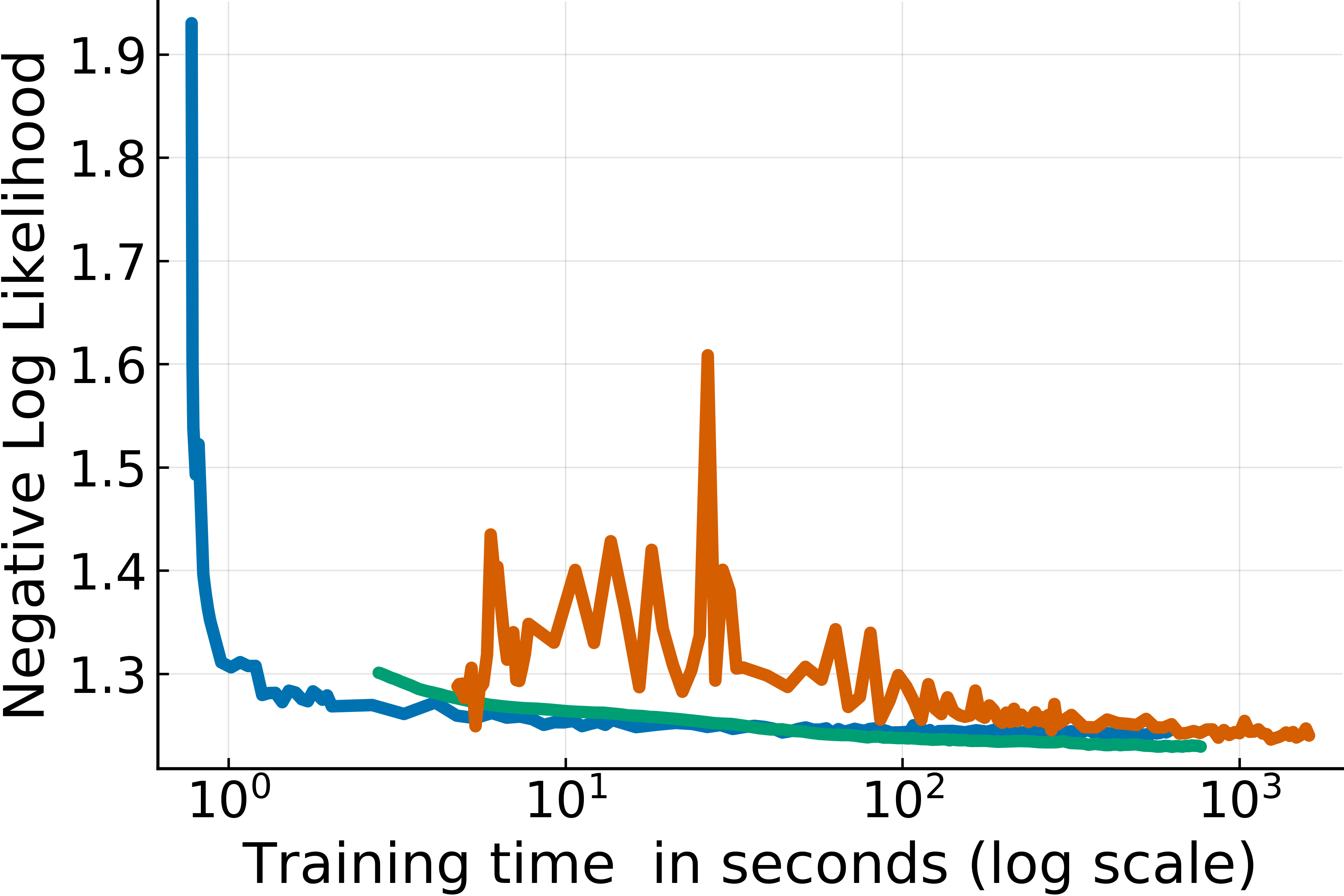}
	\includegraphics[width=0.4\textwidth]{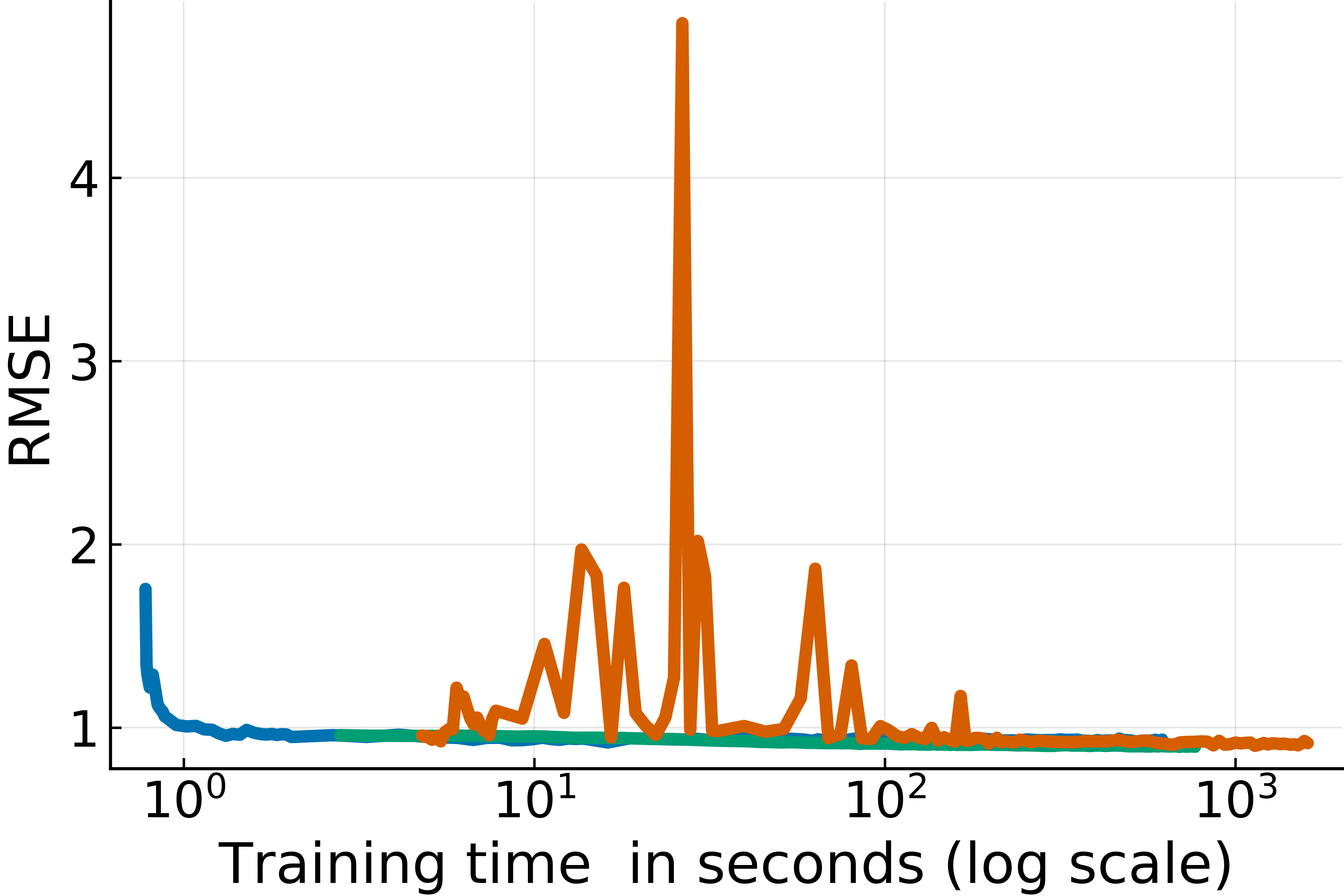}\\
	b) Matern 3/2 likelihood on the Airline dataset\\
%	\includegraphics[width=0.4\textwidth]{plots/part_3/NLL_file_name=airline_likelihood=Laplace_nInducing=200_nMinibatch=100_x_axis=time}
%	\includegraphics[width=0.4\textwidth]{plots/part_3/METRIC_file_name=airline_likelihood=Laplace_nInducing=200_nMinibatch=100_x_axis=time}\\
%	c) Laplace likelihood on the Airline dataset\\
	\includegraphics[width=0.4\textwidth]{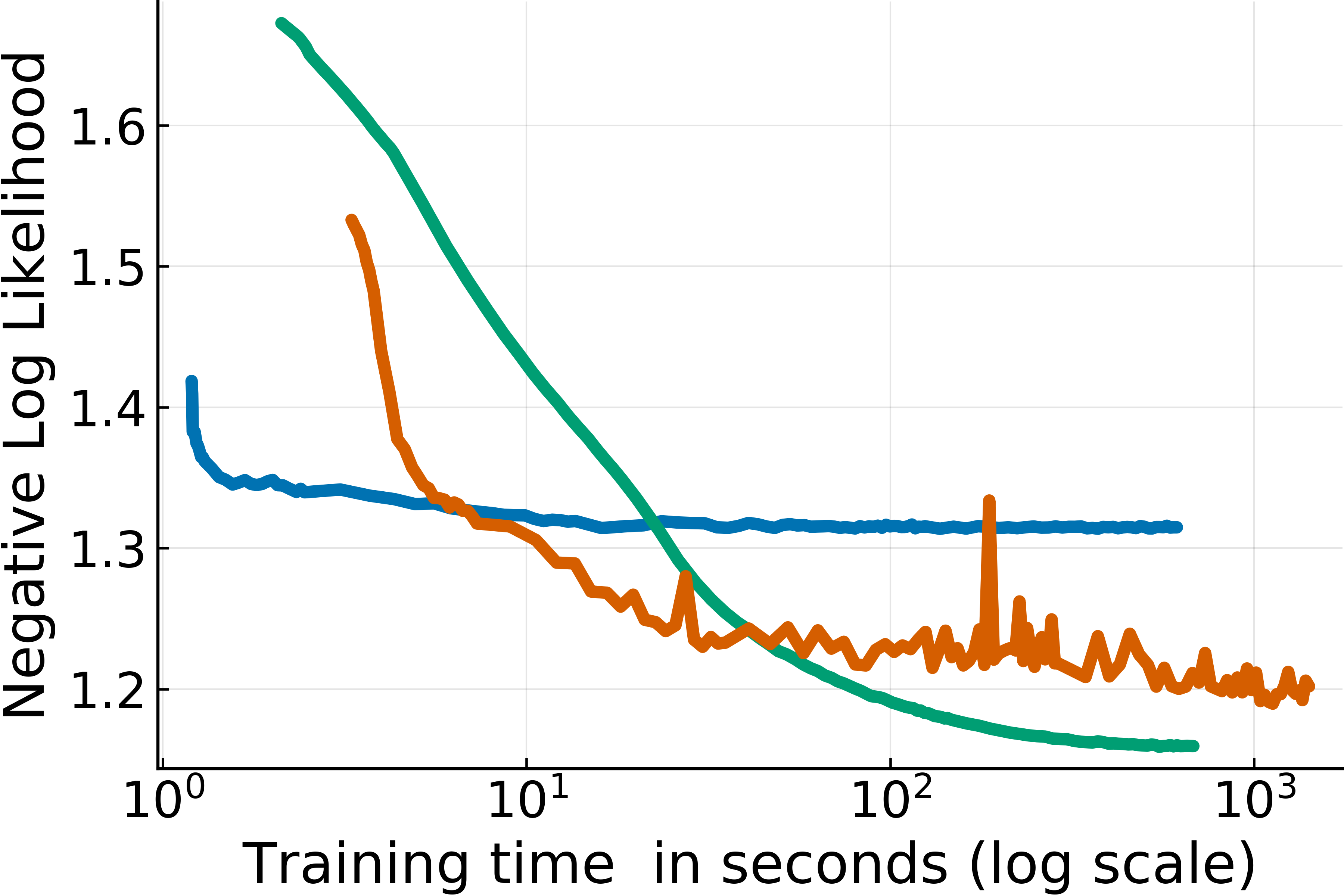}
	\includegraphics[width=0.4\textwidth]{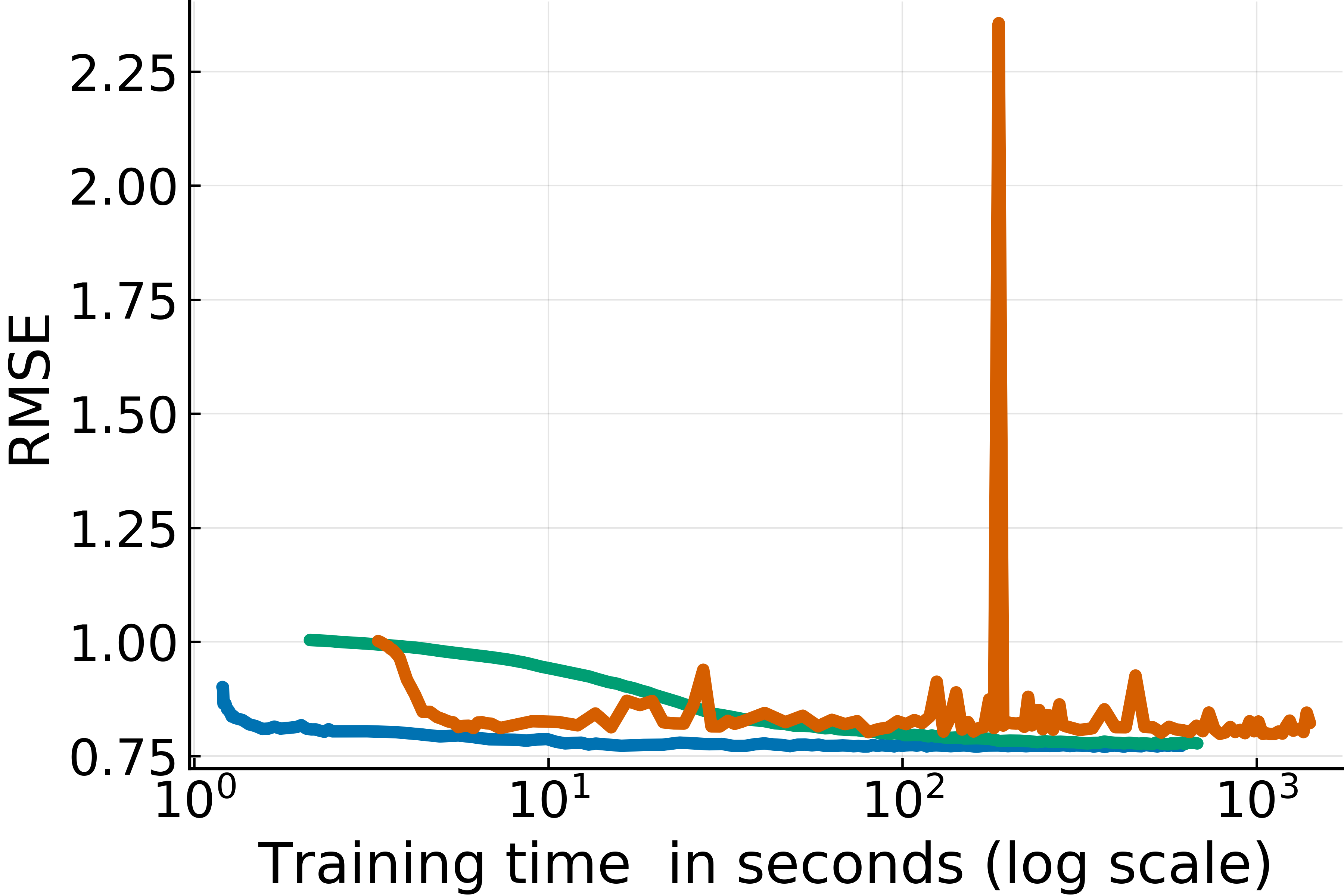}\\
	c) Student-T likelihood on the Protein dataset
\end{center}
\caption{Supplementary convergence plots}
\end{figure}

\end{document}